\documentclass{article}

\usepackage[utf8]{inputenc}
\usepackage[T1]{fontenc}
\usepackage{hyperref}
\usepackage{url}
\usepackage{booktabs}
\usepackage{amsfonts}
\usepackage{nicefrac}
\usepackage{xcolor}
\usepackage{graphicx}
\usepackage{caption}
\usepackage{subcaption}
\usepackage[margin=1in, headheight=14.5pt]{geometry}
\usepackage{verbatim}
\usepackage{setspace}
\usepackage{natbib}
\usepackage{makecell}

\title{Statistical Uncertainty Quantification for Aggregate Performance Metrics in Machine Learning Benchmarks \\\large{LA-UR-24-25289}\\ \large{Presented at the Workshop on Statistical Frontiers in LLMs and Foundation Models at NeurIPS 2024}}

\author{
  Rachel Longjohn\thanks{Denotes equal contribution.} \\
 University of California Irvine \\
  \and
  Giri Gopalan\footnotemark[1]  \\
  Los Alamos National Laboratory \\
  \and
  Emily Casleton \\
  Los Alamos National Laboratory\\
}

\begin{document}

\maketitle

\begin{abstract}
Modern artificial intelligence is supported by machine learning models (e.g., foundation models) that are pretrained on a massive data corpus and then adapted to solve a variety of downstream tasks. To summarize performance across multiple tasks, evaluation metrics are often aggregated into a summary metric, e.g., average accuracy across 10 question-answering tasks. When aggregating evaluation metrics, it is useful to incorporate uncertainty in the aggregate metric in order to gain a more realistic understanding of model performance. Our objective in this work is to demonstrate how statistical methodology can be used for quantifying uncertainty in metrics that have been aggregated across multiple tasks. The methods we emphasize are bootstrapping, Bayesian hierarchical (i.e., multilevel) modeling, and the visualization of task weightings that consider standard errors. These techniques reveal insights such as the dominance of a specific model for certain types of tasks despite an overall poor performance. We use a popular ML benchmark, the Visual Task Adaptation Benchmark (VTAB), to demonstrate the usefulness of our approaches. 
\end{abstract}

\onehalfspacing

\section{Introduction}
When evaluating machine learning (ML) models, some leaderboards compare models using \textit{point estimates} \citep{lehmann1998theory}, such as accuracy, computed on a test benchmark. However, relying only on point estimates ignores uncertainty in the evaluation metric being used to compare models. For instance, the metric estimated for a top-ranked model may not be realistically greater than the next highest-ranked model when accounting for sampling variability of test data. Many more difficulties arise when foundation models \citep{bommasani2021opportunities} are being compared as they are meant to solve many tasks instead of being trained for a single one, and to avoid making invalid conclusions about model behavior, uncertainty must be incorporated when aggregating metrics across tasks the models are adapted to solve. Our goal is to demonstrate the use of statistical methodology for aggregating metrics over many tasks that pretrained ML models are adapted to solve, which will help practitioners gain actionable insights about such models. One such insight, for instance, is that a particular model that is ranked poorly overall can be dominant when certain tasks are up-weighted.

Given the widespread adoption of ML for many critical tasks in a multitude of domains, it will be increasingly important for practitioners (e.g., individuals, companies, organizations, and research labs) to objectively evaluate pretrained models.  We demonstrate the value of statistical analysis for uncovering insights with the Visual Task Adaptation Benchmark (VTAB) \citep{zhai2019large}, an image-based benchmark that compares pretrained models that are finetuned to solve a battery of specific tasks, generally classification or question-answering using the images. 

\paragraph{Why VTAB?} There are a few advantages to using VTAB. The first is that VTAB explicitly prohibits the use of test evaluation instances in pretraining, which is not guaranteed for foundation model benchmarks (e.g., \cite{achiam2023gpt}). In addition, for VTAB, the same data are used to adapt the pretrained models being compared, such that this part of the adaptation mechanism is standardized. Finally, VTAB results are explicitly broken down by task category, which is necessary for demonstrating our visualization approaches in which we examine the role of different task weightings. More details about VTAB are summarized in the Appendix. 

\subsection{Related Work}
While the use of uncertainty quantification for task-aggregate metrics for foundation models appears to be a nascent area, the statistics and ML literature has long considered uncertainty quantification when comparing entities on scoreboards. One example is the contribution of \cite{goldstein1996league}, in which the use of bootstrapped confidence intervals and Bayesian hierarchical modeling is discussed for \textit{league tables} that rank, for example, surgeons in New York. It is shown that when incorporating uncertainty, there is less difference between the ranked surgeons than would otherwise be apparent. We take a similar approach of using both bootstrapping and Bayesian hierarchical modeling of benchmark data in order to incorporate uncertainties when comparing models on leaderboards based on benchmarks. In ML, \cite{dehghani2021benchmark} advocate for the use of statistical methods for ameliorating some of the problems with typical benchmarking practices, and \cite{colombo2022best} emphasize the use of rank aggregation across tasks via Kemeny consensus, based on minimizing average Kendall distance between the consensus rank and list of ranks. The ML literature has also suggested statistical testing and resampling procedures for evaluation, especially in natural language processing, including \cite{wein2023follow, dror2020statistical, graham2014randomized, demvsar2006statistical}. Additional relevant ML literature on rigorously comparing algorithms includes \cite{dietterich1998approximate, salzberg1997comparing, jansen2023statistical}. In contrast to these works, our work centers around statistical methods for comparing \textit{aggregate} task performance metrics for pretrained models, such as foundation models.

\subsection{Issues in Evaluating Pretrained Models and Contributions}
\begin{figure}
    \centering
    \includegraphics[width=0.8\linewidth]{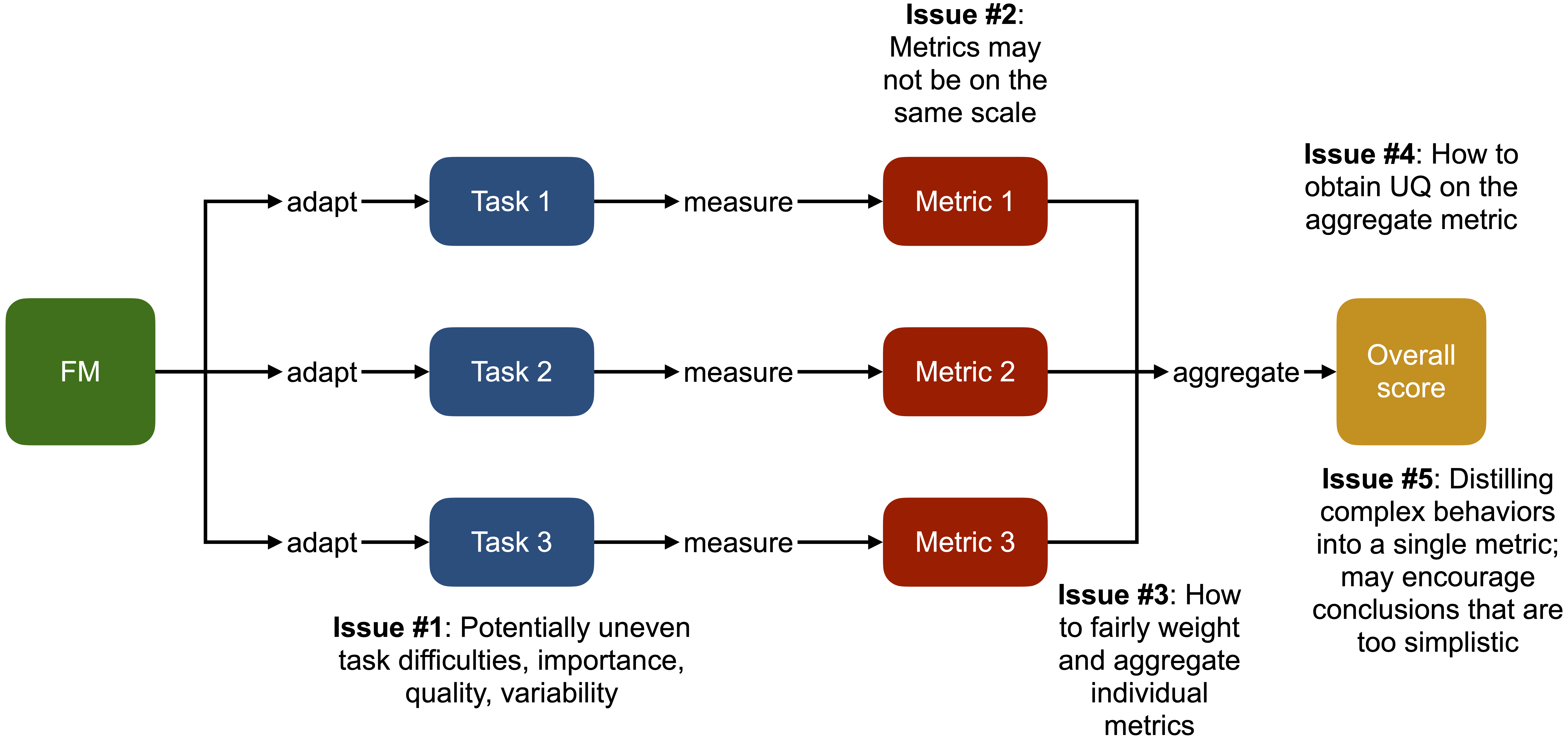}
    \caption{Illustration of challenges when evaluating foundation models.}
    \label{fig:fm-challenges}
\end{figure}

The evaluation and benchmarking of foundation models is complicated by a number of complex issues (Figure \ref{fig:fm-challenges}). Benchmark tasks show considerable variety in quality, difficulty, and importance to the end user, and so different task weightings should be examined. Further, metrics for distinct tasks may be on different scales, which must be accounted for when aggregating metrics. Finally, quantification of uncertainties for aggregated metrics must be presented in order to help accurately determine to what extent model performances differ.  A broader issue is whether ML model performance should be distilled into a single summary statistic in the first place, but we suspend argument on this point in favor of accepting that comparisons will be made based on aggregated metrics, and therefore it is pertinent to employ statistical uncertainty quantification for aggregated metrics. 

In summary, our novel contributions in this paper are:
\begin{enumerate}
\item To demonstrate how to aggregate task metrics in an ML benchmark with well-established statistical methods (bootstrapping and Bayesian hierarchical modeling) that allow for the quantification of uncertainty, and to delineate the advantages and disadvantages of using either method;
\item New visualization tools that allow for the comparison of model performance across different categories of tasks while taking uncertainty into account with standard errors, leading to more nuanced comparisons of models that convey the impact of different task weightings; and
\item To use real benchmark data from VTAB \citep{zhai2019large} to show how these statistical tools can uncover crucial insights that are important for users of pretrained ML models, including foundation models.
\end{enumerate}

\section{Statistical Methods}
We now explain the main statistical methods that we use for aggregating performance metrics across tasks. All have been used for statistical analyses in many contexts, but here we describe their use for benchmarking pretrained models that are evaluated on several tasks. We use $\theta_{ij} \in \mathbb{R}$ to denote a statistical parameter of interest that is a \textit{performance metric} for model $i$ on task $j$ for $j = 1,2,...,J$. $\theta_{ij}$ is the main quantity of interest for evaluating model $i$ on task $j$, and these $\theta_{ij}$'s may be aggregated over different tasks.

Taking a statistical viewpoint, in general the problem of aggregating task performances is equivalent to using benchmark data to estimate parameters that are functions of $\theta_{ij}$. These parameters are, for instance, the unweighted average performance across tasks for model $i$; $\bar{\theta}_{i.} := \frac{1}{J}\sum_j \theta_{ij}$; weighted aggregate performance across tasks for model $i$: $\sum_j w_{j}\theta_{ij}$; and difference in aggregate performance between model $A$ and model $B$: $\bar{\theta}_{A.}-\bar{\theta}_{B.}$.

We do not have direct access to $\theta_{ij}$ but instead can only estimate it based on benchmark data which is a random variable, so there will be uncertainty in our estimates of $\theta_{ij}$ and functions thereof. In general, these data need not be independent and could have correlation, which will be addressed. We show below several approaches to aggregate uncertainty quantification.

\subsection{Bootstrapping Evaluation Data}\label{sec:bootstrap}

\begin{figure}[!ht]
    \centering
    \includegraphics[width=0.8\linewidth]{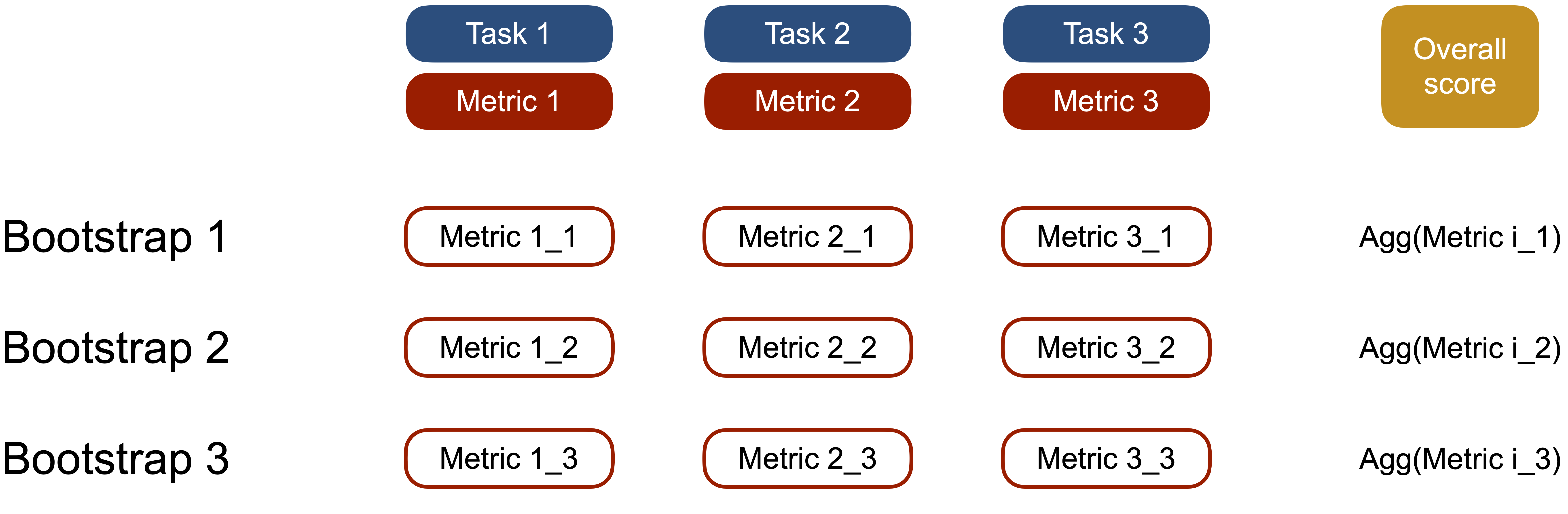}
    \caption{Illustration of bootstrap procedure for aggregate metrics.}
    \label{fig:bootstrap-viz}
\end{figure}

The first approach we employ for quantifying uncertainty in aggregate metrics is bootstrapping \citep{efron1994introduction} the test data. To generate a bootstrap sample, we resample $N_j$ test instances with replacement, where $N_j$ is the original number of instances in task $j$. The aggregate metric for a particular bootstrap sample is calculated using the benchmark's aggregation mechanism (Figure \ref{fig:bootstrap-viz}). Since VTAB takes an average across accuracies, the sample proportion of test instances correctly answered for each task is recorded as the accuracy for a given bootstrap sample, and a mean across tasks is the aggregated score for a particular model -- just as is done with the original VTAB benchmark. Uses of bootstrapping for different ML evaluation applications besides task aggregation can be found, for example, in \cite{graham2014randomized}.

From these bootstrapped aggregated scores, we derive a mean point estimate and use the appropriate quantiles to derive a confidence interval for $\bar{\theta}_{i.}$; for instance, the .025 and .975 bootstrap sample quantiles of aggregated scores for a 95\% confidence interval. We may also take differences in average performances across bootstrapped samples in order to derive confidence intervals for the difference in mean performances for models $A$ and $B$: $\bar{\theta}_{A.}-\bar{\theta}_{B.}$. We further discuss correlations between models and tasks in the subsequent subsections. 

Bootstrapping test evaluation performance is conditional on the pretrained and adapted model performances, so we are not assessing variability due to model fitting and adaptation. This is a reasonable assumption since we are interested in evaluating only fitted models (i.e., pretrained and adapted) that are to be deployed for specific tasks, as opposed to evaluating general model-fitting algorithms. Additionally, bootstrapping or employing a method such as cross validation over pretraining and training data is computationally prohibitive for foundation models, which can take on the order of many weeks to fit a single time even with large-scale computational resources (for example, see \cite{touvron2023llama}). In contrast, see \cite{bouthillier2021accounting} for an assessment of variation due to training (e.g., learning procedure and hyperparameter optimization).

When comparing confidence intervals for mean task performances across different models, it is important not to interpret overlapping 95\% confidence intervals as evidence of no statistically significant difference at $\alpha = .05$. To mitigate the possibility of this misinterpretation in our VTAB analysis, we display 83.4\% confidence intervals for mean performances, as suggested by \cite{goldstein1995graphical}; this shorter length allows overlap to correspond to an $\alpha = .05$. The more appropriate way to address comparisons of multiple models, however, is to determine confidence intervals for pairwise differences in aggregated task performance. We also present these confidence intervals along with a multiple comparisons Bonferroni correction, which guards against a single false rejection of the null amongst several hypotheses but can be conservative for many comparisons. This is not much of a concern for us, however, because we simultaneously compare only the top few models.

The bootstrap samples also provide a means to estimate ranks with confidence intervals for models' aggregate performance for various rank aggregation mechanisms. For example, one may use the sample ranks based on average task performance for each bootstrap sample in order to estimate the overall model ranks, where the population model ranks are defined by ranking the average performance parameters $\bar{\theta}_{i.}$. 

\subsection{Bayesian Hierarchical Modeling of Evaluation Data}\label{sec:BHM}
An alternative to the bootstrap for quantifying uncertainties in task-aggregate metrics is a Bayesian hierarchical model (BHM). In comparison to bootstrapping, this approach can be advantageous because a distribution over each model's task performances can encode prior knowledge of model performance, helping to adjust estimates of model performance to be more realistic and less variable when some tasks have much fewer items compared to others.  Also one can explicitly model correlations between different task performances or between models; for example, between-task correlations would not be accounted for due to the resampling procedure discussed previously for the bootstrap.

On the other hand, prior information about model performance can be difficult to elicit and encode into a prior distribution, and fitting a BHM can be more difficult computationally than bootstrapping. Nonetheless, a BHM is another important option for quantifying uncertainties in aggregate task metrics. We note that Bayesian modeling has been used in ML evaluation before. For example, in \cite{ji2021active}, they use beta-binomial models in a sequential design setting and also suggest the potential utility of a hierarchical approach. In \cite{benavoli2017time}, they employ Bayesian hierarchical modeling for the problem of modeling cross-validation data for classifiers.

Here we illustrate a beta-binomial hierarchical model for the analysis of benchmarks that use accuracy as the performance metric. 
For model $i$ and task $j$, we denote $Y_{ij}$ as the number of correct responses, $N_j$ as the number of test items for task $j$, and $\theta_{ij}$ as the probability of a correct response for model $i$ on task $j$. Further, we assume a binomial model for the number of correct responses along with a beta (conjugate) prior for each task-model probability. That is:

$$Y_{ij} | \theta_{ij} \sim Binom(\theta_{ij}, N_j)$$ is the data model, and the model for $\theta_{ij}$ is
$$\theta_{ij} | \alpha_i, \beta_i \sim Beta(\alpha_i, \beta_i).$$
Conditioning on $\theta_{ij}$, $Y_{ij}$ is independent of the rest of the data, and likewise for parameter $\theta_{ij}$ conditional on $\alpha_i, \beta_i$. 
The BHM specification is completed by putting a distribution on $\alpha_i, \beta_i$:
$$\alpha_i \sim Exp(\lambda_{\alpha})$$ and $$\beta_i \sim Exp(\lambda_{\beta}).$$

These are assumed to be independent draws $\forall i$. An important variation of the above model is when the data for each task-model combination are in the form of confusion matrices --  widely used metrics like F1-score are derived from these counts. To accommodate this setting, one could modify the binomial distribution to be a multinomial, where the probabilities are of true positive, true negative, false positive, and false negative and a Dirichlet prior is used; see \cite{totsch2021classifier} for the multinomial Dirichlet set up. However, using a hierarchy as above (i.e., a prior for the Dirichlet parameters) allows for shrinkage of estimates and for the incorporation of prior knowledge of model performance.

The posterior of the above beta-binomial hierarchical model is not available in closed form. We thus implemented a Gibbs sampler for VTAB in which the conditional distribution of $\theta_{ij}$ given all else is a beta distribution by conjugacy. However, the conditional distributions of $\alpha_i$ given all else and $\beta_i$ given all else are not analytically recognizable distributions and we use a slice sampling MCMC step within Gibbs \citep{neal2003slice}. One could instead use a probabilistic programming language to sample if desired. We use $\lambda_{\alpha} = \lambda_{\beta} = 1/10000$ to allow for a wide range of plausible prior success and failure counts; however, to increase shrinkage, one could use a more stringent prior, for instance that shrinks estimates of $\theta_{ij}$ to be closer to the grand mean of performance of model $i$. To illustrate, we conducted a basic simulation study as follows. 

\paragraph{Simulation study} Consider when two models are compared on three question-and-answering tasks. Let $\theta_{ij}$ denote model $i$'s probability of accurate response on task $j$, and assume that $\theta_{Aj} = .5$ for $j = 1,2,3$ while $\theta_{B1} = .525$ and $\theta_{B2} = \theta_{B3} = .5$; \textit{hence model B is the better performing model}. Let the test set sizes be $N_1 = 200$ for the first task, $N_2 = 10000$ for the second task, and $N_3 = 20000$ for the third task. Assume that after evaluating the two models on the three tasks, we obtain the number of correct responses in Table \ref{tab:sim-data}.

\begin{table}[!ht]
\centering
\caption{Summary of the number of correct questions answered by two models in three question-and-answering tasks.}
\begin{tabular}{c c c c} 
\toprule
 & \makecell{\textbf{Task 1}\\ $N_1 = 200$} & \makecell{\textbf{Task 2}\\ $N_2 = 10000$} & \makecell{\textbf{Task 3}\\ $N_3 = 20000$} \\
\midrule
 Model A & 100 & 5000 & 10000 \\ 
 Model B & 115 & 5000 & 10000 \\
\bottomrule
\end{tabular}
\label{tab:sim-data}
\end{table}

We consider the parameter of interest $\bar{\theta}_{A.}-\bar{\theta}_{B.}$ for the difference in mean performances between models A and B, which we know is truly negative in this example. A 95\% bootstrapped confidence interval that uses the difference in average sample accuracies across the three tasks yields the following: (-.059, .0097); since the interval contains 0, one arrives at the wrong conclusion that there is no difference between the model performances. 

In comparison, consider instead a beta-binomial Bayesian hierarchical model as previously described that allows us to encode prior information on task performances, which dampens the effect of increased variability because of a small number of questions in task 1. Specifically, consider normal priors for the prior success and failure counts $\alpha_i$ and $\beta_i$:
\begin{itemize}
\item $\alpha_1 \sim N(2000, 10)$
\item $\alpha_2 \sim N(2100, 10)$
\item $\beta_1  \sim N(2000, 10)$
\item $\beta_2  \sim N(1900, 10)$,
\end{itemize}
where $N(\mu,\sigma)$ indicates a normal distribution with mean $\mu$ and standard deviation $\sigma$; we further truncate at 0 to ensure positivity, though the difference is negligible since 0 is about 200 standard deviations below the mean. These priors adjust estimates for tasks that have a small number of questions (task 1 in this example) such that they have less variability (due to 4000 prior counts). Here, we find that the 95\% credibility interval for $\bar{\theta}_{A.}-\bar{\theta}_{B.}$ is (-0.021, -0.003). Since the interval is strictly negative, we are led to the correct conclusion that model B is better than model A. 

Once posterior samples of $\theta_{ij}$ are drawn, one can sample from the posterior predictive distribution of task performances in order to obtain a posterior predictive credibility interval for aggregated accuracy and also derive posterior rank probabilities based on ranks of $\sum_{j=1}^J w_j\theta_{ij}$, a weighted sum of task performance parameters; more on the analysis of task weightings is provided in the next section. 

\subsection{Weighting Task Performances}\label{sec:weighting}
We consider weighting task performances with weights $w_1, ..., w_J$ such that $\sum_{j=1}^J w_j = 1$ and $w_j \geq 0$. Let $Y_{Aj}$ and $Y_{Bj}$ be task performances of model $A$ and model $B$ respectively for $j = 1, ...,J$. Model $A$'s aggregated score is $S_A := \sum_{j=1}^J w_jY_{Aj}$ and model $B$'s aggregated score is $S_B := \sum_{j=1}^J w_jY_{Bj}$; if task performances are unbiased estimators of performance parameters $\theta_{Aj}$ and $\theta_{Bj}$, then by linearity of expectation, $S_A$ and $S_B$ are unbiased for the aggregate performance parameters $\sum_{j=1}^J w_j \theta_{Aj}$ and $\sum_{j=1}^J w_j \theta_{Bj}$, respectively. However, concluding that the aggregate performance of model $A$ is greater than that of model $B$ because $S_A > S_B$ is incorrect because it ignores the variances of $S_A$ and $S_B$. 

For example, the variance of $S_A$ is calculated by:
\begin{eqnarray}
                Var[S_A] &=& Var\left[\sum_{j = 1}^Jw_jY_{Aj}\right] \\
                         &=& \sum_{j=1}^Jw_j^2Var[Y_{Aj}] + 2\sum_{j < j'}w_jw_{j'}Cov[Y_{Aj}, Y_{Aj'}].
\end{eqnarray}
Additional derivations for standard errors specific to VTAB are provided in the Appendix. We note that one consequence of this identity is that if task performances are positively correlated, then the variance of $S_A$ assuming independence lower bounds the variance of $S_A$ with dependence. 

An additional observation will help to interpret the task-weighting visualizations that follow when considering correlations between the performances of two models. The plots depend on the variance of the difference in scores between the top two performing models:
\begin{eqnarray}
Var[S_A - S_B] &=& Var[S_A] + Var[S_B] - 2\rho_{A,B} \sqrt{Var[S_A]Var[S_B]},
\end{eqnarray}
where $\rho_{A,B}$ is the correlation between aggregated task performance of model $A$ and model $B$. We can use this identity to investigate the effect of correlation between top-performing models in the visualizations we use in our VTAB analysis. For instance, if the variances of $S_A$ and $S_B$ are close, and $\rho = .5$, then the standard deviation of the difference is about 70\% of the value in the independent case. The Appendix contains a more general identity for addressing between-model correlation. 

\subsection{Metric Normalization}

An approach to normalizing metrics is given by BigBench \citep{srivastava2022beyond}, which is $\frac{raw-low}{high-low}$, where $high$ is an upper bound for a task score, $low$ is a lower bound for a task score, and $raw$ is the unadjusted score on a task -- such normalization maps scores to the interval [0, 1]. One approach to $high$ and $low$ scores is to use a pre-defined baseline score (e.g., a baseline for accuracy could correspond to guessing randomly). However this can be more difficult to determine for a high score, and for the low score, one may want to consider floors that are better than the worst possible scenario. In the following analysis, we estimate the high and low scores from  bootstrap samples; $high$ for a particular task is the maximum score over all bootstrap samples, and similarly for $low$.

An alternative approach to putting all tasks on the same scale is to use rankings, for instance as is suggested by \cite{colombo2022best}. One drawback to rank aggregation is that it ignores information in the distances between different models in the process of transforming data to a list of ranks. Nonetheless, aggregating rankings is an important approach to compare models, which we illustrate in a variety of ways in the following section. There is also a growing body of statistical literature on performing ranking with quantified uncertainties -- see for example \cite{goldstein1996league, deng2014bayesian, xie2009confidence, rising2021uncertainty, barrientos2023bayesian, li2022bayesian, hall2010modeling}.

\section{Application to VTAB}

We applied the techniques described above to evaluate the performance of all 16 models from the VTAB 1-k leaderboard. For ease of presentation, we focus on the results for the top six models here and make the complete results available in the Appendix. We also include a summary of the VTAB benchmark in the Appendix. Since we do not have the per-task responses across the 16 fine-tuned models compared for VTAB, we used simulated data consistent with the per-task accuracies for each model on the VTAB leaderboard \footnote{\url{https://google-research.github.io/task_adaptation/benchmark}}, and we extract the per-task test set sizes from the VTAB GitHub repository \footnote{\url{https://github.com/google-research/task_adaptation}}. This application may serve as an example of the value of applying the statistical methods described previously to aggregation of task metrics for pretrained models in general benchmarks.

\subsection{Confidence and Credibility Intervals for Aggregated Performances}
Table \ref{tab:aggregate-accs} shows the point and interval estimates for each model's unweighted aggregate performance on the benchmark using: 1) across-task average accuracy with 83.4\% confidence intervals using the bootstrapped test data, 2) across-task average accuracy with 83.4\% credible intervals using the BHM, and 3) across-task average normalized accuracy with 83.4\% confidence intervals using the bootstrapped test data. Recall we display 83.4\% intervals as based on \cite{goldstein1995graphical}, as to mitigate misinterpretation of overlapping 95 \% confidence intervals. Overall, the results in Table \ref{tab:aggregate-accs} suggest that the two supervised methods, Sup-Rotation-100\% (SR-100\%) and Sup-Exemplar-100\% (SE-100\%), perform the best but that there are not meaningful differences in these two models’ unweighted aggregate performance. As expected, the point estimates for the unnormalized average accuracy preserve the original ranking of the models, with SR-100\% having the highest estimated average accuracy and SE-100\% in second place. Looking at this alone might suggest SR-100\%’s dominance. When looking at the results using the average normalized accuracy, the possible conclusions change; the ranking of the top two models according to their point estimates is reversed, and their confidence intervals are non-overlapping. Doing well under the normalization scheme suggests that a model performs particularly strong in harder tasks, in which the minimum observed value of the performance metric is lower. Under the normalization scheme, the results suggest that Sup-Exemplar-100\% would be the dominant model. Similar findings could be expressed when comparing the semi-supervised versions of these two models as well.

\begin{table}[!ht]
    \centering
    \caption{Average accuracy across VTAB's 19 tasks with interval estimates using the bootstrapped evaluation data and the Bayesian hierarchical model, and average normalized accuracy using the bootstrapped evaluation data. We display 83.4\% intervals, as per \cite{goldstein1995graphical}.}
    \begin{tabular}{lccc}
    \toprule
    \textbf{Model} & \makecell{\textbf{Avg Acc}\\\textbf{(Bootstrap)}} & \makecell{\textbf{Avg Acc}\\\textbf{(Bayesian HM)}} & \textbf{Avg Norm Acc} \\
    \midrule
    Sup-Rotation-100\% & 68.0 (67.8, 68.1) & 68.0 (67.7, 68.2) & 88.4 (87.9, 88.7) \\
    Sup-Exemplar-100\% & 67.6 (67.4, 67.9) & 67.6 (67.4, 67.9) & 89.5 (89.0, 90.0) \\
    Sup-100\% & 66.4 (66.2, 66.5) & 66.3 (66.1, 66.6) & 85.1 (84.6, 85.6) \\
    Semi-Exemplar-10\% & 65.3 (65.1, 65.5) & 65.3 (65.0, 65.6) & 83.3 (82.9, 83.6) \\
    Semi-Rotation-10\% & 65.1 (64.9, 65.3) & 65.1 (64.8, 65.3) & 83.7 (83.3, 84.1)\\
    Rotation & 60.4 (60.3, 60.6) & 60.4 (60.2, 60.7) & 77.6 (77.3, 78.0)\\
    \bottomrule
    \end{tabular}
    \label{tab:aggregate-accs}
\end{table}

We investigate this further by conducting pairwise comparisons in the performance differences of the top three models. As discussed in Section \ref{sec:bootstrap}, when comparing two models, the pairwise difference in mean task performance is the true quantity of interest. Table \ref{tab:top3-pairwise} presents the point and interval estimates for these pairwise differences, using the Bonferroni correction for conducting three comparisons. Across all three metric uncertainty quantification techniques, it cannot be concluded that there are differences in the unweighted aggregate performance of SR-100\% and SE-100\%; both models outperform Sup-100\% (S-100\%).

\begin{table}[!ht]
    \centering
    \caption{Point estimates and confidence intervals for the pairwise differences in average accuracies and average normalized accuracies for the top three performing models. We display 95\% intervals with a Bonferroni adjustment for 3 comparisons.}
    \begin{tabular}{lccc}
    \toprule
    \textbf{Comparison} & \makecell{\textbf{Avg Acc Diff}\\\textbf{(Bootstrap)}} & \makecell{\textbf{Avg Acc Diff}\\\textbf{(Bayesian HM)}}& \textbf{Avg Norm Acc Diff}\\
    \midrule
    SR-100\% - SE-100\% & 0.3 (-0.1, 0.8) & 0.3 (-0.3, 0.9) & -1.1 (-2.1, 0.0) \\
    SR-100\% - S-100\% & 1.6 ( 1.2, 2.0) & 1.6 ( 1.0, 2.2) & { }3.3 ( 2.3, 4.3) \\
    SE-100\% - S-100\% & 1.3 ( 0.9, 1.7)&  1.3 ( 0.7, 1.9) & { }4.4 ( 3.4, 5.3)\\
    \bottomrule
    \end{tabular}
    \label{tab:top3-pairwise}
\end{table}

\subsection{Confidence and Credible Intervals for Aggregated Ranks}
In Tables \ref{tab:rank-uq-top6} and \ref{tab:norm-rank-uq-top6}, we present 95\% confidence intervals for several rank aggregation schemes for the unnormalized and normalized metrics, respectively. This includes several rank aggregation techniques based on those used in the appendix of the VTAB paper: 1) ranking by the unweighted arithmetic mean of the accuracies, 2) ranking by the geometric mean, 3) rank first on a per-task basis and then take the across-task average rank, 4) same as 3 but add in standard normal noise before ranking, and 5) same as 3 but bin the accuracies into buckets of size 1\% before ranking, where being in the same bin results in a tie rank. We can also use samples from the posterior predictive distribution of the BHM to get credible intervals, which give similar results (see Table \ref{tab:bayes-rank-uq} in the Appendix). Across all rank aggregation strategies, we see general agreement with the conclusions above. Within the supervised and semi-supervised model classes, there do not appear to be meaningful differences in overall performance between the Rotation and Exemplar versions of these models. Interestingly, there is some disagreement across the rank aggregation strategies about the relative performance of the Rotation model, originally placed sixth. The aggregation strategies based on the average rank can reward more heavily slight edges in performance across multiple tasks while only lightly penalizing large gaps in performance that do not change relative positions much. This suggests that there are several tasks for which Rotation performs competitively, or outperforms, previously higher-ranking models.

\begin{table}[!ht]
\footnotesize
    \centering
    \caption{95\% confidence intervals for rank aggregation of unnormalized accuracy using bootstrapped test data.}
    \begin{tabular}{lccccc}
    \toprule
    \textbf{Model} & \textbf{By avg} & \textbf{Geom mean} & \textbf{Avg rank (AvR)} & \textbf{AvR (noise)} & \textbf{AvR (bins)} \\
    \midrule
    SR-100\% & 1.0 (1.0,  2.0) & 1.4 (1.0,  2.0) & 3.8 (3.6,  4.0) & 3.8 (3.2,  4.4) & 4.2 (3.9, 4.4)\\
    SE-100\% & 2.0 (1.0,  2.0) & 1.6 (1.0,  2.0) & 3.9 (3.6,  4.2) & 4.0 (3.5,  4.6) & 4.5 (4.1,  4.8)\\
    S-100\% & 3.0 (3.0,  3.0) & 3.0 (3.0,  3.0) & 5.0 (4.7,  5.3) & 5.0 (4.5,  5.7) & 5.5 (5.1,  5.9)\\
    Semi-E-10\% & 4.1 (4.0,  5.0) & 4.0 (4.0,  4.5) & 5.5 (5.3,  5.7) & 5.4 (4.8,  5.9) & 5.9 (5.6,  6.2)\\
    Semi-R-10\% & 4.9 (4.0,  5.0) & 5.0 (4.5,  5.0) & 5.4 (5.1,  5.7) & 5.3 (4.7,  5.9) & 5.9 (5.5,  6.2)\\
    Rotation & 6.0 (6.0,  6.0) & 6.0 (6.0,  6.0) & 4.9 (4.7,  5.2) & 5.0 (4.4,  5.4) & 5.2 (5.0, 5.4)\\
    \bottomrule
    \end{tabular}
    \label{tab:rank-uq-top6}
\end{table}

\begin{table}[!ht]
\footnotesize
    \centering
    \caption{95\% confidence intervals for rank aggregation of normalized accuracy using bootstrapped test data.} 
\begin{tabular}{lccccc}
\toprule
\textbf{Model} & \textbf{By avg} & \textbf{Geom mean} & \textbf{Avg rank (AvR)} & \textbf{AvR (noise)} & \textbf{AvR (bins)} \\
\midrule
SR-100\% & 2.0 (2.0,  2.0) & 2.0 (2.0,  2.0) & 3.8 (3.6,  4.0) & 3.9 (3.5,  4.2) & 3.9 (3.7,  4.2)\\
SE-100\% & 1.0 (1.0,  1.0) & 1.0 (1.0,  1.0) & 3.9 (3.6,  4.2) & 3.9 (3.6,  4.3) & 4.1 (3.8,  4.4)\\
S-100\% & 3.0 (3.0,  3.0) & 3.0 (3.0,  3.0) & 5.0 (4.7,  5.3) & 5.0 (4.5,  5.4) & 5.2 (4.8,  5.6)\\
Semi-E-10\% & 4.8 (4.0,  5.0) & 4.9 (4.0,  5.0) & 5.5 (5.3,  5.7) & 5.5 (5.1,  5.8) & 5.6 (5.4,  5.8)\\
Semi-R-10\% & 4.2 (4.0,  5.0) & 4.1 (4.0,  5.0) & 5.4 (5.1,  5.7) & 5.3 (5.0,  5.7) & 5.6 (5.3,  5.9)\\
Rotation & 6.0 (6.0,  6.0) & 6.0 (6.0,  6.0) & 4.9 (4.7,  5.2) & 5.0 (4.7,  5.2) & 5.0 (4.8,  5.3)\\
\bottomrule
\end{tabular}
    \label{tab:norm-rank-uq-top6}
\end{table}

\subsection{Visualizing Task Weightings with Uncertainty}

In some application settings, it may make sense to weigh particular tasks more heavily in the evaluation, e.g., if a practitioner cares more about a model doing well at tasks with structured images but would still like the model to perform reasonably well at tasks with natural or specialized images. In such instances, it may be difficult to specify or justify a particular instantiation of such weights that makes sense (e.g., 0.7, 0.15, 0.15), but instead one may only be able to express some qualitative intuitions, such as caring most about structured images. 

To visualize multiple weighting schemes simultaneously, we construct a plot on the simplex which shades a region according to which model performs the best for particular weighting schemes in which the weights sum to 1. These plots were created using plotting functions from the open-source R package, Ternary \citep{Smith2017}, with customizations to account for discrete-valued shading. Gray regions on the plot indicate an indeterminate result, in which the differences in performance for the top two performing models is within $z$ standard errors of 0. Here, a standard error is computed for the difference in weighted task performance between the top two models. We focus on weighting across the three different task categories in VTAB: natural, specialized, and structured images.  

\begin{figure}[!ht]
    \caption{Visualization of differences in model performance under different category weightings using the unnormalized accuracies.}
    \label{fig:weight-simplex}
    \begin{subfigure}[t]{0.5\textwidth}
        \centering
        \includegraphics[width=\textwidth]{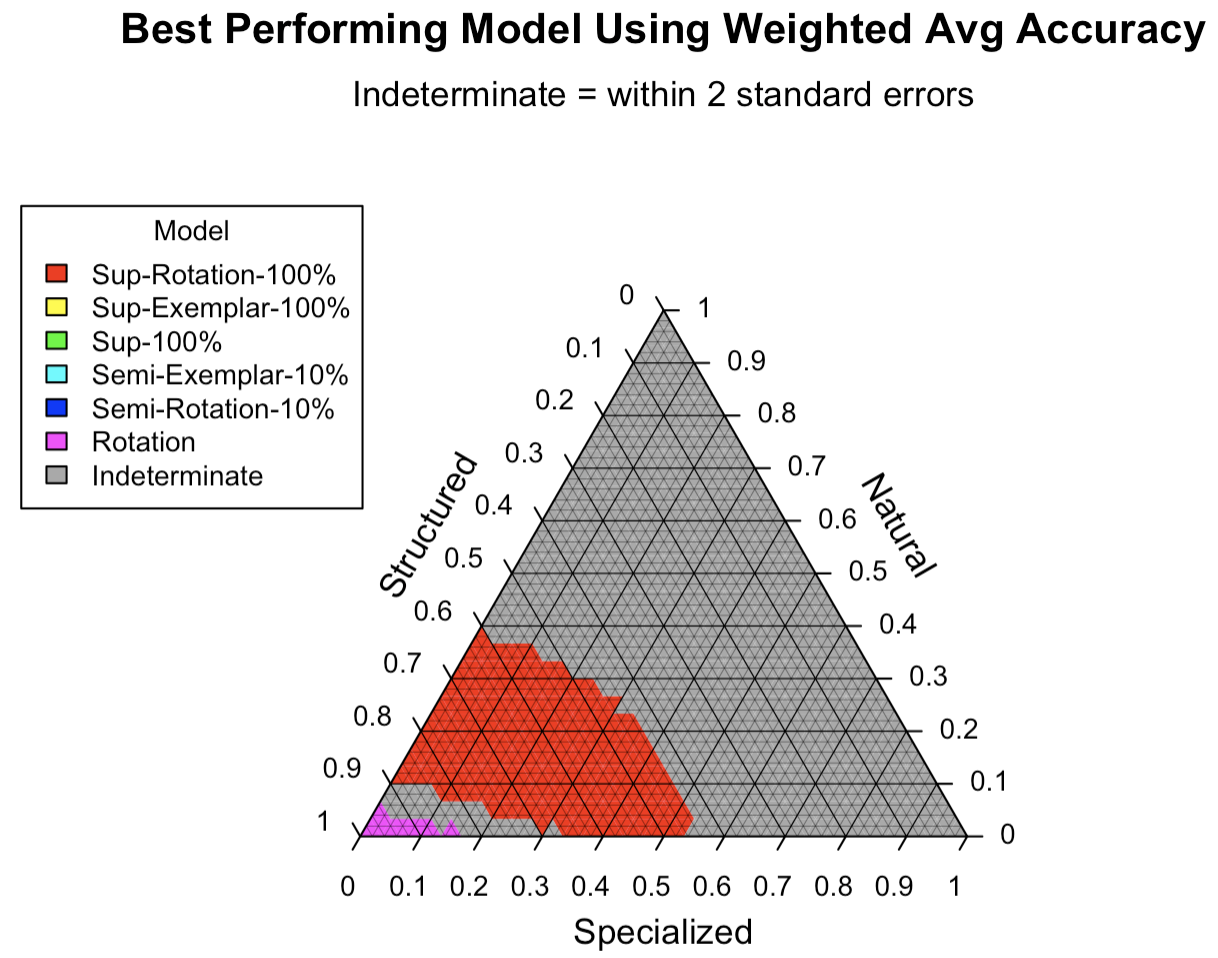}
        \caption{Independence (2 standard errors).}
    \end{subfigure}
    \hfill
    \begin{subfigure}[t]{0.5\textwidth}
        \centering
        \includegraphics[width=\textwidth]{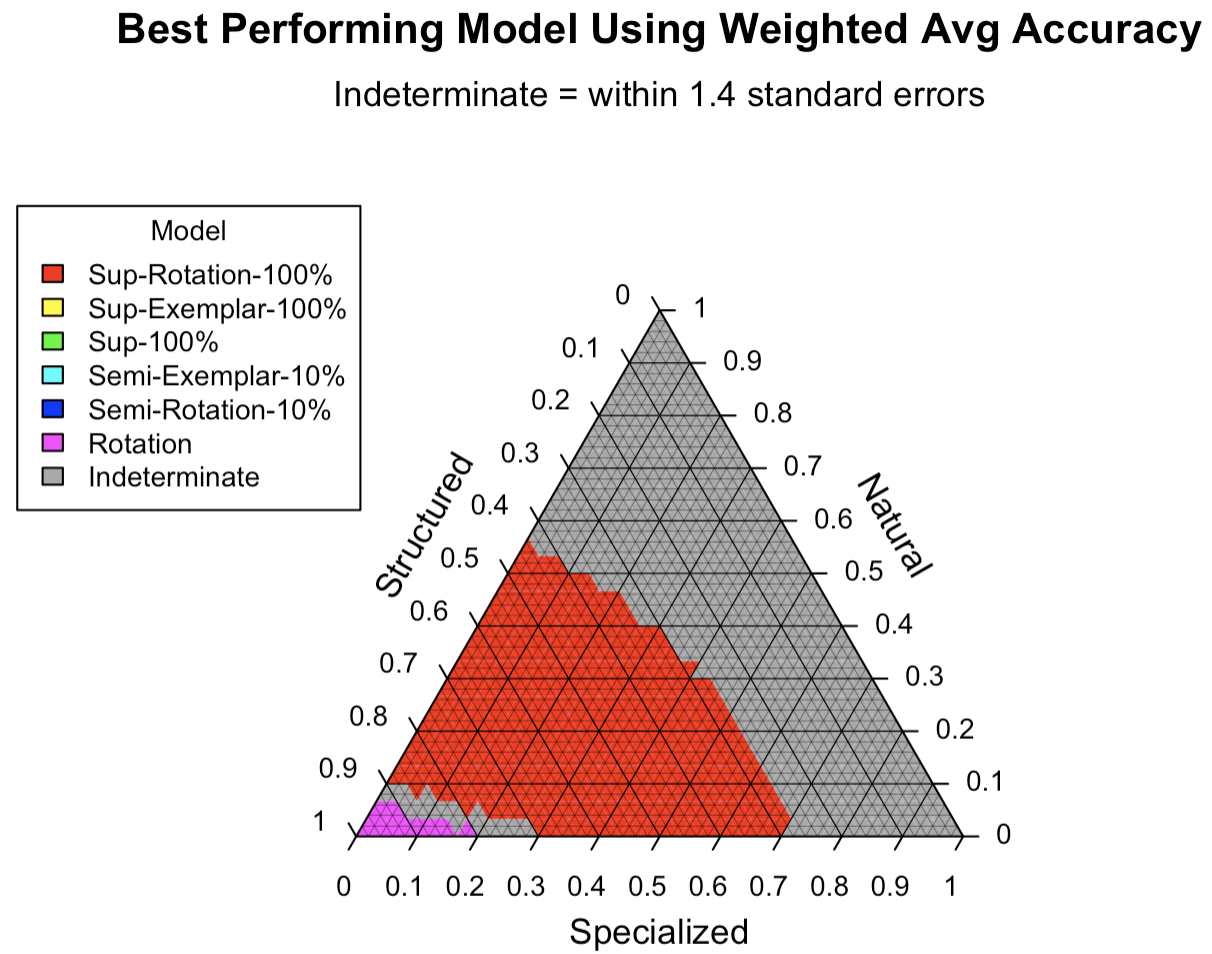}
        \caption{Dependence (1.4 standard errors).}
    \end{subfigure}
\end{figure}

\begin{figure}[!ht]
    \caption{Visualization of differences in model performance under different category weightings using the normalized accuracies.}
    \label{fig:norm-weight-simplex}
    \begin{subfigure}[t]{0.5\textwidth}
        \centering
        \includegraphics[width=\textwidth]{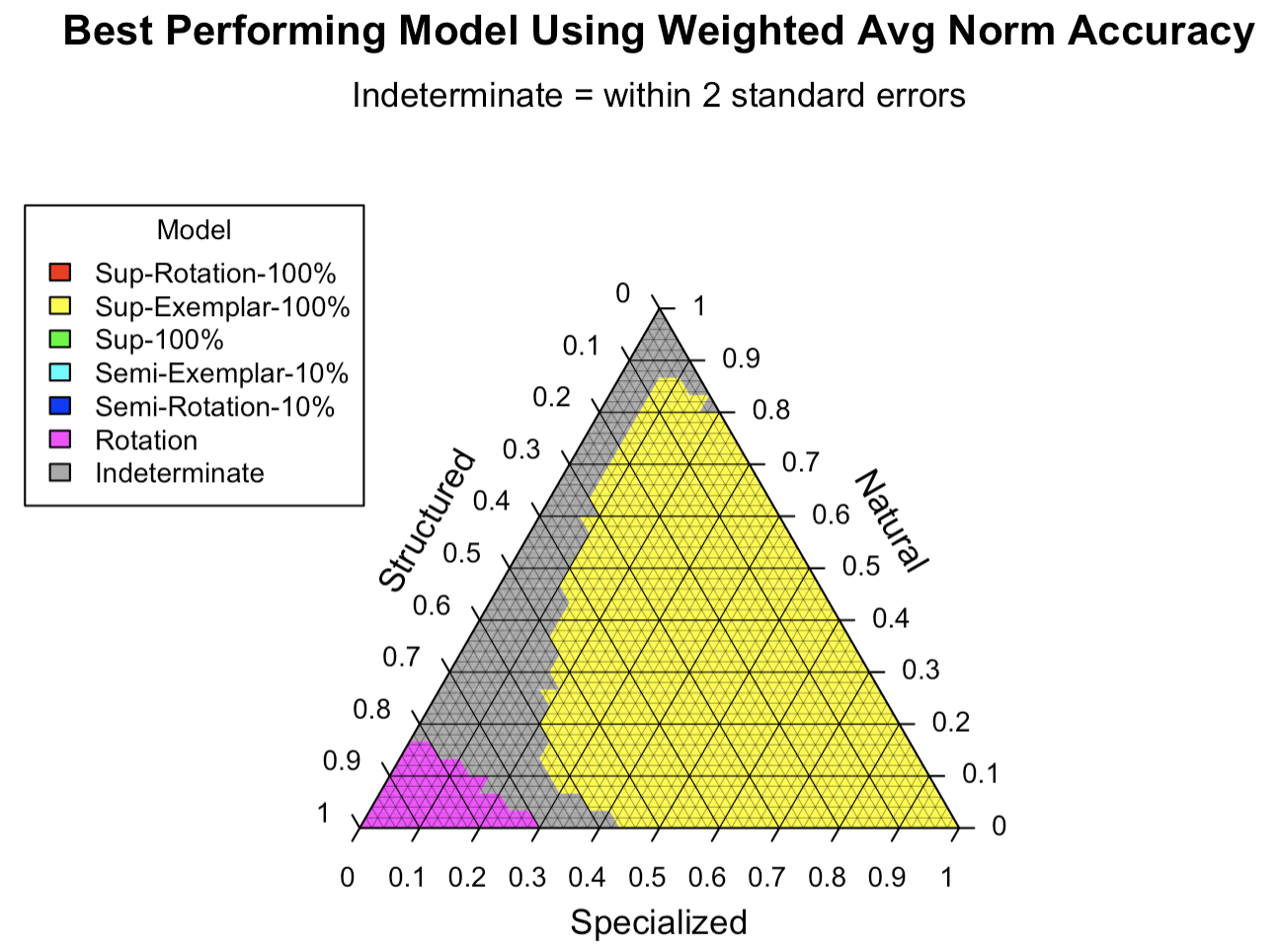}
        \caption{Independence (2 standard errors).}
    \end{subfigure}
    \hfill
    \begin{subfigure}[t]{0.5\textwidth}
        \centering
        \includegraphics[width=\textwidth]{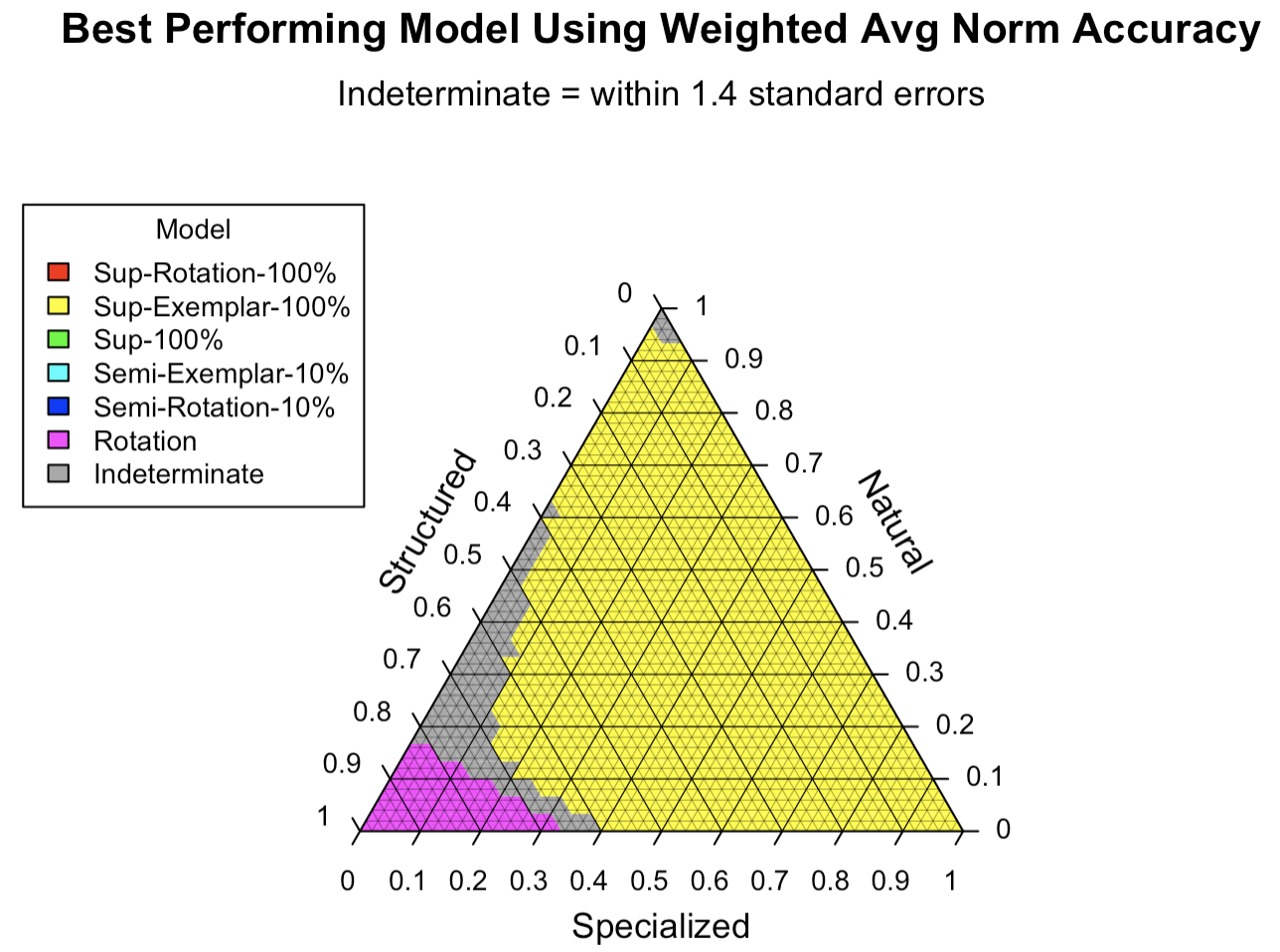}
        \caption{Dependence (1.4 standard errors).}
    \end{subfigure}
\end{figure}

Figures \ref{fig:weight-simplex} and \ref{fig:norm-weight-simplex} show these plots for the unnormalized and normalized metrics, respectively. Based on the findings in Section \ref{sec:weighting} related to the potential correlations between model performance, we show these plots for values of $z = 2$ and $z=2\sqrt{1/2} \approx 1.4$ to account for the potential reduction in standard deviation that could be present for correlated model performances. A robust result across all of these plots is that when structured images are weighted heavily in the aggregation, Rotation is suggested as the best performing model. This aligns with the findings suggested by the average rank-based aggregation schemes in Tables \ref{tab:rank-uq-top6} and \ref{tab:norm-rank-uq-top6}. Similar conclusions are also suggested by the BHM (Figure \ref{fig:weighted-bayes-ranks} in the Appendix).

Taken together, the differences between Figures \ref{fig:weight-simplex} and \ref{fig:norm-weight-simplex} further reinforce that there do not appear to be meaningful differences in performance between the SR-100\% and SE-100\% models for many of the possible values of the weights. Unlike in the pairwise differences in Table \ref{tab:top3-pairwise}, the values in these plots are not meant to indicate statistical significance but instead are intended to be a graphical depiction which still takes into account uncertainty. Such visualizations could be a helpful tool in settings in which model decisions need to be made, or model evaluations need to be communicated to relevant stakeholders.

\section{Discussion and Conclusion}

Our main contribution is to demonstrate statistical methods for aggregating task performances of pretrained and adapted models, such as foundation models, with quantified uncertainties. We have used benchmark data from VTAB in performing this demonstration and have highlighted bootstrapping of task evaluation data, Bayesian hierarchical modeling, and visualizing task weighting with standard errors. The value of this type of statistical analysis was demonstrated by revealing insights such as the fact that a low-ranked model (Rotation) is superior depending on task weighting (while accounting for uncertainty). An additional insight is that certain models may essentially be of the same overall performance when uncertainty is accounted for, despite that their aggregate point estimates are distinct. These statistical analyses will ideally serve as a blueprint for the benchmarking of foundation models for practitioners, and we aim to develop ways to incorporate prediction uncertainty in future work.

\newpage

\section*{Acknowledgments}
The authors would like to thank the NNSA (NA-22) Office of Defense Nuclear Nonproliferation Research and Development for supporting this research and Padhraic Smyth and Natalie Klein for feedback on earlier versions of this paper. The LA-UR for this work is \textit{LA-UR-24-25289}.
\newpage

\bibliographystyle{plainnat}
\newpage
\bibliography{res}

\newpage

\clearpage
\appendix

\section{Appendix}

\subsection{VTAB Benchmark Summary}
We demonstrate statistical methods based on evaluation data from a widely-used, existing benchmark: the visual task adaptation benchmark (VTAB) \citep{zhai2019large}. VTAB is used for image (e.g., computer vision) models, originally developed to assess representation learning. The benchmark consists of 19 classification tasks split into 3 categories: natural, specialized, and structured. We use data on the 16 pretrained models compared on the original VTAB leaderboard \footnote{\url{https://google-research.github.io/task_adaptation/benchmark}}, though VTAB has been used for benchmarking additional models. The natural category consists of tasks  such as classifying pictures of animals; specialized tasks include medical problems such as predicting the presence of metastatic cancerous tissue; structured tasks include predicting the orientation of an object in an image and finding the depth of an object in an image. For this work we rely on VTAB-1k, which allows each model 1000 labeled images per task to use in the adaptation stage, but does not constrain resources allowed for pretraining. There are between 711 and 73,728 test instances in the 19 tasks provided by VTAB, the numbers of which can be verified at the GitHub repository \footnote{\url{https://github.com/google-research/task_adaptation}}. Models are scored on VTAB by aggregating per-task accuracy with an unweighted average across the 19 tasks. VTAB aggregates accuracies on distinct tasks for leaderboards. VTAB is a prime benchmark to use for our statistical analyses because it standardizes the data resources allowed for adaptation of a pretrained model, and ``\textit{representations are not pretrained on the evaluation tasks themselves (which VTAB forbids)}'' \citep{zhai2019large}. We note that VTAB does standardize the data resources allowed for training and validation when adapting pretrained vision models compared; moreover VTAB also shows evidence that fine-tuning stabilizes after a fixed number of steps.

\subsection{Standard Error Derivations}
Standard error calculations for VTAB task weighting plots are given here assuming independence between tasks; Section \ref{sec:weighting} addresses the issue of correlation between tasks. Let $\hat{p}_{Aj}$ be the accuracy of model $A$ on natural task $j$ of $N_{Nat}$. Let $\bar{p}_{A,Nat}$ be average accuracy for the natural tasks for model $A$, which is:
\begin{eqnarray}
\bar{p}_{A,Nat} &:=& N_{Nat}^{-1}\sum_{j=1}^{N_{Nat}} \hat{p}_{Aj}
\end{eqnarray}
We have thus the following variance.
\begin{eqnarray}
Var[\bar{p}_{A,Nat}] &=&  N_{Nat}^{-2}\sum_{j=1}^{N_{Nat}} Var[\hat{p}_{Aj}].
\end{eqnarray}
We use a binomial standard error squared for $Var[\hat{p}_{Aj}]$, which is $\frac{\hat{p}_{Aj}(1-\hat{p}_{Aj})}{N_j}$.
The same holds for the specialized and structured categories. Now, we consider the variance of the weighted score for model $A$:
\begin{eqnarray}
S_A &:=& w_1\bar{p}_{A,Sp} + w_2\bar{p}_{A,Str}+ w_3\bar{p}_{A,Nat}
\end{eqnarray}
From which the variance can be calculated:
\begin{eqnarray}
Var[S_A] &=& w_1^2Var[\bar{p}_{A,Sp}] + w_2^2Var[\bar{p}_{A,Str}]+ w^2_3Var[\bar{p}_{A,Nat}].
\end{eqnarray}

The standard errors used in the task weighting simplex plots are calculated based on $Var[S_A - S_B]$, which is $Var[S_A] + Var[S_B]$, if independence is assumed between distinct models. Section \ref{sec:weighting} addresses how we account for correlation between models by reducing the standard error; here we include a more general statement:

Denote $Var[S_A] := \sigma^2$ and write  $Var[S_B] := k\sigma^2$ for some positive constant $k \geq 1$ (assume without loss of generality that $Var[S_B] \geq Var[S_A]$). Then, $Var[S_A - S_B] = \sigma^2 +k\sigma^2 - 2\rho\sqrt{(k\sigma^2)\sigma^2}$. Thus, the standard deviation of the difference assuming correlation is reduced by a factor of $\sqrt{\frac{(k+1)-2\rho\sqrt{k}}{k+1}}$. 

\subsection{Full Results}
\begin{table}[!ht]
\footnotesize
    \centering
    \caption{VTAB leaderboard with 83.4\% interval estimates using the bootstrapped standard errors.}
    \begin{tabular}{lllll}
    \toprule
    \textbf{Model} & \textbf{Natural} & \textbf{Specialized} & \textbf{Structured} & \textbf{Overall}\\
    \midrule
    Sup-Rotation-100\% & 73.6 (73.2, 73.9) & 83.1 (82.9, 83.3) & 55.5 (55.2, 55.7) & 68.0 (67.8, 68.1)\\
    Sup-Exemplar-100\% & 73.6 (73.3, 74.0) & 83.1 (82.8, 83.4) & 54.7 (54.3, 55.0) & 67.6 (67.4, 67.9)\\
    Sup-100\% & 73.4 (73.2, 73.7) & 82.5 (82.2, 82.7) & 52.1 (51.7, 52.5) & 66.4 (66.2, 66.5)\\
    Semi-Exemplar-10\% & 70.2 (70.0, 70.5) & 81.8 (81.6, 82.1) & 52.7 (52.4, 53.1) & 65.3 (65.1, 65.5)\\
    Semi-Rotation-10\% & 69.5 (69.2, 69.8) & 82.4 (82.2, 82.6) & 52.5 (52.2, 52.9) & 65.1 (64.9, 65.3)\\
    \addlinespace
    Rotation & 53.7 (53.3, 54.0) & 78.6 (78.3, 78.8) & 57.3 (57.0, 57.6) & 60.4 (60.3, 60.6)\\
    Exemplar & 48.9 (48.6, 49.2) & 78.4 (78.2, 78.6) & 55.8 (55.4, 56.2) & 58.0 (57.8, 58.3)\\
    Rel.Pat.Loc & 46.0 (45.7, 46.3) & 76.5 (76.3, 76.7) & 48.3 (48.0, 48.6) & 53.4 (53.2, 53.6)\\
    Jigsaw & 44.0 (43.7, 44.3) & 76.5 (76.3, 76.9) & 47.9 (47.6, 48.2) & 52.5 (52.3, 52.7)\\
    Uncond-BigGAN & 35.9 (35.6, 36.2) & 63.0 (62.7, 63.4) & 45.7 (45.4, 46.0) & 45.8 (45.6, 46.0)\\
    \addlinespace
    From-Scratch & 27.9 (27.6, 28.2) & 68.9 (68.6, 69.2) & 43.6 (43.2, 43.9) & 43.1 (43.0, 43.3)\\
    Cond-BigGAN & 39.5 (39.2, 39.8) & 57.4 (57.0, 57.8) & 35.1 (34.7, 35.4) & 41.4 (41.2, 41.6)\\
    WAE-MMD & 20.8 (20.6, 21.0) & 60.6 (60.4, 60.9) & 43.4 (43.1, 43.7) & 38.7 (38.5, 38.8)\\
    VAE & 19.4 (19.2, 19.6) & 59.2 (58.9, 59.5) & 44.2 (43.9, 44.6) & 38.2 (38.0, 38.4)\\
    WAE-UKL & 15.0 (14.8, 15.2) & 55.2 (55.0, 55.6) & 39.0 (38.7, 39.3) & 33.6 (33.4, 33.8)\\
    WAE-GAN & 15.6 (15.4, 15.8) & 54.0 (53.8, 54.2) & 38.5 (38.1, 38.8) & 33.3 (33.1, 33.5)\\
    \bottomrule
    \end{tabular}
    \label{tab:leaderboard-conf}
\end{table}

\begin{table}[!ht]
\footnotesize
    \centering
    \caption{VTAB leaderboard with 83.4\% credible intervals from the Bayesian hierarchical model.}
    \begin{tabular}{lllll}
    \toprule
    Model & Natural & Specialized & Structured & Overall\\
    \midrule
    Sup-Rotation-100\% & 73.6 (73.2, 74.0) & 83.1 (82.8, 83.4) & 55.5 (55.0, 55.9) & 68.0 (67.7, 68.2)\\
    Sup-Exemplar-100\% & 73.6 (73.2, 74.1) & 83.1 (82.7, 83.4) & 54.7 (54.2, 55.1) & 67.6 (67.4, 67.9)\\
    Sup-100\% & 73.4 (73.0, 73.8) & 82.5 (82.1, 82.8) & 52.1 (51.6, 52.6) & 66.3 (66.1, 66.6)\\
    Semi-Exemplar-10\% & 70.2 (69.8, 70.7) & 81.8 (81.5, 82.2) & 52.7 (52.2, 53.2) & 65.3 (65.0, 65.5)\\
    Semi-Rotation-10\% & 69.5 (69.1, 70.0) & 82.4 (82.0, 82.7) & 52.5 (52.0, 53.0) & 65.1 (64.8, 65.3)\\
    \addlinespace
    Rotation & 53.7 (53.2, 54.2) & 78.5 (78.2, 78.9) & 57.3 (56.8, 57.7) & 60.4 (60.2, 60.7)\\
    Exemplar & 48.9 (48.5, 49.4) & 78.4 (78.0, 78.7) & 55.8 (55.3, 56.2) & 58.0 (57.7, 58.3)\\
    Rel.Pat.Loc & 46.0 (45.5, 46.5) & 76.5 (76.2, 76.9) & 48.4 (47.9, 48.9) & 53.4 (53.2, 53.7)\\
    Jigsaw & 44.0 (43.5, 44.4) & 76.5 (76.1, 76.9) & 47.9 (47.4, 48.3) & 52.5 (52.2, 52.7)\\
    Uncond-BigGAN & 35.9 (35.5, 36.4) & 63.1 (62.6, 63.5) & 45.7 (45.2, 46.2) & 45.8 (45.5, 46.0)\\
    \addlinespace
    From-Scratch & 27.9 (27.5, 28.3) & 68.9 (68.5, 69.4) & 43.6 (43.0, 44.1) & 43.1 (42.9, 43.4)\\
    Cond-BigGAN & 39.5 (39.0, 40.0) & 57.4 (56.9, 57.9) & 35.1 (34.6, 35.6) & 41.4 (41.1, 41.7)\\
    WAE-MMD & 20.8 (20.5, 21.1) & 60.6 (60.2, 61.0) & 43.4 (42.9, 43.9) & 38.7 (38.4, 38.9)\\
    VAE & 19.4 (19.1, 19.8) & 59.2 (58.8, 59.6) & 44.2 (43.7, 44.7) & 38.2 (38.0, 38.5)\\
    WAE-UKL & 15.1 (14.8, 15.4) & 55.2 (54.8, 55.7) & 39.0 (38.5, 39.5) & 33.6 (33.3, 33.9)\\
    WAE-GAN & 15.6 (15.3, 15.9) & 54.0 (53.5, 54.4) & 38.4 (37.9, 38.9) & 33.3 (33.0, 33.5)\\
    \bottomrule
    \end{tabular}
    \label{tab:bayes-leaderboard}
\end{table}

\begin{table}[!ht]
\scriptsize
    \centering
    \caption{Rank aggregation of average accuracy using bootstrapped test data.}
\begin{tabular}{llllll}
\toprule
 \textbf{Model} & \textbf{By avg} & \textbf{Geom mean} & \textbf{Avg rank (AvR)} & \textbf{AvR (noise)} & \textbf{AvR (bins)}\\
\midrule
SR-100\% & 1.0 ( 1.0,  2.0) & 1.4 ( 1.0,  2.0) & 3.8 ( 3.6,  4.0) & 3.8 ( 3.2,  4.4) & 4.2 ( 3.9,  4.4)\\
SE-100\% & 2.0 ( 1.0,  2.0) & 1.6 ( 1.0,  2.0) & 3.9 ( 3.6,  4.2) & 4.0 ( 3.5,  4.6) & 4.5 ( 4.1,  4.8)\\
Sup-100\% & 3.0 ( 3.0,  3.0) & 3.0 ( 3.0,  3.0) & 5.0 ( 4.7,  5.3) & 5.0 ( 4.5,  5.7) & 5.5 ( 5.1,  5.9)\\
Semi-E-10\% & 4.1 ( 4.0,  5.0) & 4.0 ( 4.0,  4.5) & 5.5 ( 5.3,  5.7) & 5.4 ( 4.8,  5.9) & 5.9 ( 5.6,  6.2)\\
Semi-R-10\% & 4.9 ( 4.0,  5.0) & 5.0 ( 4.5,  5.0) & 5.4 ( 5.1,  5.7) & 5.3 ( 4.7,  5.9) & 5.9 ( 5.5,  6.2)\\
\addlinespace
Rotation & 6.0 ( 6.0,  6.0) & 6.0 ( 6.0,  6.0) & 4.9 ( 4.7,  5.2) & 5.0 ( 4.4,  5.4) & 5.2 ( 5.0,  5.4)\\
Exemplar & 7.0 ( 7.0,  7.0) & 7.0 ( 7.0,  7.0) & 6.1 ( 5.8,  6.4) & 6.0 ( 5.4,  6.4) & 6.4 ( 6.2,  6.7)\\
Rel.Pat.Loc & 8.0 ( 8.0,  8.0) & 8.0 ( 8.0,  8.0) & 8.5 ( 8.4,  8.7) & 8.4 ( 8.0,  8.9) & 8.7 ( 8.5,  8.9)\\
Jigsaw & 9.0 ( 9.0,  9.0) & 9.0 ( 9.0,  9.0) & 9.2 ( 8.9,  9.4) & 9.1 ( 8.6,  9.6) & 9.5 ( 9.2,  9.8)\\
Uncond-BigGAN & 10.0 (10.0, 10.0) & 10.0 (10.0, 10.0) & 10.3 (10.1, 10.5) & 10.3 ( 9.9, 10.7) & 10.5 (10.2, 10.7)\\
\addlinespace
From-Scratch & 11.0 (11.0, 11.0) & 12.0 (12.0, 12.0) & 10.9 (10.6, 11.3) & 10.9 (10.5, 11.4) & 11.3 (11.0, 11.5)\\
Cond-BigGAN & 12.0 (12.0, 12.0) & 11.0 (11.0, 11.0) & 10.6 (10.5, 10.8) & 10.8 (10.4, 11.3) & 10.9 (10.7, 11.1)\\
WAE-MMD & 13.0 (13.0, 13.0) & 13.6 (13.0, 14.0) & 11.8 (11.6, 12.0) & 11.8 (11.4, 12.3) & 12.1 (11.9, 12.4)\\
VAE & 14.0 (14.0, 14.0) & 13.4 (13.0, 14.0) & 11.7 (11.5, 11.9) & 11.8 (11.4, 12.2) & 11.9 (11.8, 12.1)\\
WAE-UKL & 15.1 (15.0, 16.0) & 15.0 (15.0, 15.0) & 14.0 (13.7, 14.3) & 14.0 (13.6, 14.4) & 14.4 (14.2, 14.6)\\
WAE-GAN & 15.9 (15.0, 16.0) & 16.0 (16.0, 16.0) & 14.4 (14.2, 14.7) & 14.3 (13.9, 14.8) & 14.7 (14.5, 14.9)\\
\bottomrule
\end{tabular}
    \label{tab:rank-uq}
\end{table}

\begin{table}[!ht]
\footnotesize
    \centering
    \caption{VTAB leaderboard with 83.4\% interval estimates for the normalized accuracies using the bootstrapped standard errors.}
    \begin{tabular}{lllll}
    \toprule
    \textbf{Model} & \textbf{Natural} & \textbf{Specialized} & \textbf{Structured} & \textbf{Overall}\\
    \midrule
    Sup-Rotation-100\% & 95.6 (95.1, 96.1) & 89.2 (88.3, 90.2) & 81.5 (80.9, 82.2) & 88.4 (87.9, 88.7)\\
    Sup-Exemplar-100\% & 96.3 (95.8, 96.9) & 92.7 (91.8, 93.9) & 81.8 (81.0, 82.7) & 89.5 (89.0, 90.0)\\
    Sup-100\% & 96.0 (95.5, 96.4) & 90.2 (89.2, 91.3) & 73.0 (72.0, 74.0) & 85.1 (84.6, 85.6)\\
    Semi-Exemplar-10\% & 90.0 (89.6, 90.4) & 88.7 (87.7, 89.5) & 74.7 (73.9, 75.6) & 83.3 (82.9, 83.6)\\
    Semi-Rotation-10\% & 88.1 (87.6, 88.6) & 90.6 (89.7, 91.6) & 76.4 (75.6, 77.3) & 83.7 (83.3, 84.1)\\
    \addlinespace
    Rotation & 62.2 (61.7, 62.7) & 78.6 (77.5, 79.6) & 90.6 (89.9, 91.4) & 77.6 (77.3, 78.0)\\
    Exemplar & 54.3 (53.8, 54.8) & 80.8 (79.7, 81.7) & 82.9 (81.9, 83.8) & 71.9 (71.4, 72.4)\\
    Rel.Pat.Loc & 48.7 (48.2, 49.1) & 71.6 (70.6, 72.5) & 62.4 (61.7, 63.3) & 59.3 (58.9, 59.8)\\
    Jigsaw & 45.8 (45.3, 46.3) & 74.0 (73.0, 75.2) & 60.5 (59.7, 61.4) & 57.9 (57.5, 58.3)\\
    Uncond-BigGAN & 34.8 (34.4, 35.3) & 39.0 (37.9, 40.0) & 56.1 (55.3, 56.9) & 44.7 (44.2, 45.2)\\
    \addlinespace
    From-Scratch & 22.3 (22.0, 22.8) & 63.2 (62.1, 64.3) & 48.5 (47.6, 49.4) & 42.0 (41.5, 42.5)\\
    Cond-BigGAN & 40.2 (39.8, 40.8) & 50.7 (49.5, 51.7) & 27.0 (26.1, 27.8) & 36.8 (36.3, 37.3)\\
    WAE-MMD & 12.0 (11.7, 12.3) & 50.6 (49.7, 51.7) & 44.8 (44.0, 45.4) & 33.9 (33.5, 34.3)\\
    VAE & 9.3 (9.0, 9.7) & 33.1 (32.0, 34.2) & 50.9 (50.0, 51.9) & 31.8 (31.4, 32.4)\\
    WAE-UKL & 2.0 (1.7, 2.3) & 39.0 (37.9, 40.0) & 32.0 (31.2, 32.8) & 22.4 (22.0, 22.8)\\
    WAE-GAN & 3.1 (2.9, 3.5) & 35.7 (34.7, 36.9) & 30.3 (29.4, 31.2) & 21.4 (21.0, 21.9)\\
    \bottomrule
    \end{tabular}
    \label{tab:leaderboard-conf-norm}
\end{table}

\begin{table}[!ht]
\scriptsize
    \centering
    \caption{Rank aggregation of average normalized accuracy using bootstrapped test data.}
\begin{tabular}{llllll}
\toprule
\textbf{Model} & \textbf{By avg} & \textbf{Geom mean} & \textbf{Avg rank (AvR)} & \textbf{AvR (noise)} & \textbf{AvR (bins)}\\
\midrule
SR-100\% & 2.0 ( 2.0,  2.0) & 2.0 ( 2.0,  2.0) & 3.8 ( 3.6,  4.0) & 3.9 ( 3.5,  4.2) & 3.9 ( 3.7,  4.2)\\
SE-100\% & 1.0 ( 1.0,  1.0) & 1.0 ( 1.0,  1.0) & 3.9 ( 3.6,  4.2) & 3.9 ( 3.6,  4.3) & 4.1 ( 3.8,  4.4)\\
Sup-100\% & 3.0 ( 3.0,  3.0) & 3.0 ( 3.0,  3.0) & 5.0 ( 4.7,  5.3) & 5.0 ( 4.5,  5.4) & 5.2 ( 4.8,  5.6)\\
Semi-E-10\% & 4.8 ( 4.0,  5.0) & 4.9 ( 4.0,  5.0) & 5.5 ( 5.3,  5.7) & 5.5 ( 5.1,  5.8) & 5.6 ( 5.4,  5.8)\\
Semi-R-10\% & 4.2 ( 4.0,  5.0) & 4.1 ( 4.0,  5.0) & 5.4 ( 5.1,  5.7) & 5.3 ( 5.0,  5.7) & 5.6 ( 5.3,  5.9)\\
\addlinespace
Rotation & 6.0 ( 6.0,  6.0) & 6.0 ( 6.0,  6.0) & 4.9 ( 4.7,  5.2) & 5.0 ( 4.7,  5.2) & 5.0 ( 4.8,  5.3)\\
Exemplar & 7.0 ( 7.0,  7.0) & 7.0 ( 7.0,  7.0) & 6.1 ( 5.8,  6.4) & 6.1 ( 5.7,  6.4) & 6.2 ( 6.0,  6.5)\\
Rel.Pat.Loc & 8.0 ( 8.0,  8.0) & 8.0 ( 8.0,  8.0) & 8.5 ( 8.4,  8.7) & 8.5 ( 8.3,  8.7) & 8.6 ( 8.4,  8.8)\\
Jigsaw & 9.0 ( 9.0,  9.0) & 9.0 ( 9.0,  9.0) & 9.2 ( 8.9,  9.4) & 9.2 ( 8.8,  9.5) & 9.3 ( 9.1,  9.6)\\
Uncond-BigGAN & 10.0 (10.0, 10.0) & 10.2 (10.0, 13.4) & 10.3 (10.1, 10.5) & 10.3 (10.0, 10.5) & 10.3 (10.1, 10.6)\\
\addlinespace
From-Scratch & 11.0 (11.0, 11.0) & 10.9 (10.0, 11.0) & 10.9 (10.6, 11.3) & 10.9 (10.7, 11.2) & 11.0 (10.8, 11.3)\\
Cond-BigGAN & 12.0 (12.0, 12.0) & 13.5 (12.0, 15.8) & 10.6 (10.5, 10.8) & 10.7 (10.5, 10.9) & 10.7 (10.5, 10.8)\\
WAE-MMD & 13.0 (13.0, 13.0) & 12.2 (11.5, 13.0) & 11.8 (11.6, 12.0) & 11.8 (11.5, 12.1) & 11.9 (11.7, 12.1)\\
VAE & 14.0 (14.0, 14.0) & 13.4 (12.0, 14.0) & 11.7 (11.5, 11.9) & 11.7 (11.5, 12.0) & 11.8 (11.6, 12.0)\\
WAE-UKL & 15.0 (15.0, 15.0) & 15.0 (14.0, 16.0) & 14.0 (13.7, 14.3) & 14.0 (13.7, 14.2) & 14.1 (13.8, 14.3)\\
WAE-GAN & 16.0 (16.0, 16.0) & 15.9 (15.0, 16.0) & 14.4 (14.2, 14.7) & 14.4 (14.1, 14.6) & 14.5 (14.3, 14.7)\\
\bottomrule
\end{tabular}
    \label{tab:norm-rank-uq}
\end{table}

\begin{table}[!ht]
\scriptsize
    \centering
    \caption{Rank aggregation of accuracies using samples from the BHM.}
\begin{tabular}{lccccc}
\toprule
Model & By avg & Geom mean & Avg rank (AvR) & AvR (noise) & AvR (bins)\\
\midrule
Sup-Rotation-100\% & 1.1 ( 1.0,  2.0) & 1.4 ( 1.0,  2.0) & 3.8 ( 3.5,  4.2) & 3.9 ( 3.3,  4.5) & 4.2 ( 3.8,  4.6)\\
Sup-Exemplar-100\% & 1.9 ( 1.0,  2.0) & 1.6 ( 1.0,  2.0) & 3.9 ( 3.5,  4.3) & 4.0 ( 3.4,  4.6) & 4.5 ( 4.1,  4.9)\\
Sup-100\% & 3.0 ( 3.0,  3.0) & 3.0 ( 3.0,  3.0) & 5.0 ( 4.6,  5.4) & 5.0 ( 4.4,  5.6) & 5.5 ( 5.1,  5.9)\\
Semi-Exemplar-10\% & 4.2 ( 4.0,  5.0) & 4.1 ( 4.0,  5.0) & 5.5 ( 5.2,  5.8) & 5.4 ( 4.8,  6.0) & 5.9 ( 5.6,  6.3)\\
Semi-Rotation-10\% & 4.8 ( 4.0,  5.0) & 4.9 ( 4.0,  5.0) & 5.4 ( 5.0,  5.7) & 5.3 ( 4.7,  5.9) & 5.9 ( 5.5,  6.3)\\
\addlinespace
Rotation & 6.0 ( 6.0,  6.0) & 6.0 ( 6.0,  6.0) & 4.9 ( 4.7,  5.2) & 5.0 ( 4.5,  5.4) & 5.2 ( 4.9,  5.5)\\
Exemplar & 7.0 ( 7.0,  7.0) & 7.0 ( 7.0,  7.0) & 6.1 ( 5.8,  6.4) & 6.0 ( 5.5,  6.6) & 6.4 ( 6.1,  6.7)\\
Rel.Pat.Loc & 8.0 ( 8.0,  8.0) & 8.0 ( 8.0,  8.0) & 8.5 ( 8.2,  8.7) & 8.4 ( 7.9,  8.8) & 8.7 ( 8.4,  8.9)\\
Jigsaw & 9.0 ( 9.0,  9.0) & 9.0 ( 9.0,  9.0) & 9.2 ( 8.8,  9.5) & 9.1 ( 8.5,  9.6) & 9.5 ( 9.2,  9.8)\\
Uncond-BigGAN & 10.0 (10.0, 10.0) & 10.0 (10.0, 10.0) & 10.3 (10.0, 10.5) & 10.3 ( 9.8, 10.7) & 10.5 (10.2, 10.7)\\
\addlinespace
From-Scratch & 11.0 (11.0, 11.0) & 12.0 (12.0, 12.0) & 11.0 (10.6, 11.2) & 10.9 (10.4, 11.5) & 11.3 (10.9, 11.5)\\
Cond-BigGAN & 12.0 (12.0, 12.0) & 11.0 (11.0, 11.0) & 10.7 (10.5, 10.9) & 10.8 (10.4, 11.3) & 10.9 (10.7, 11.2)\\
WAE-MMD & 13.0 (13.0, 14.0) & 13.5 (13.0, 14.0) & 11.8 (11.5, 12.1) & 11.9 (11.4, 12.4) & 12.1 (11.8, 12.5)\\
VAE & 14.0 (13.0, 14.0) & 13.5 (13.0, 14.0) & 11.7 (11.5, 11.9) & 11.8 (11.4, 12.2) & 11.9 (11.7, 12.2)\\
WAE-UKL & 15.1 (15.0, 16.0) & 15.0 (15.0, 15.0) & 14.0 (13.7, 14.3) & 14.0 (13.6, 14.5) & 14.3 (14.0, 14.6)\\
WAE-GAN & 15.9 (15.0, 16.0) & 16.0 (16.0, 16.0) & 14.4 (14.1, 14.7) & 14.3 (13.8, 14.8) & 14.7 (14.4, 15.0)\\
\bottomrule
\end{tabular}
    \label{tab:bayes-rank-uq}
\end{table}

\begin{figure}
\setlength\tabcolsep{3pt}
\centering
\begin{tabular}{ccc}
\begin{subfigure}[t]{0.3\textwidth}
    \includegraphics[width=\textwidth]{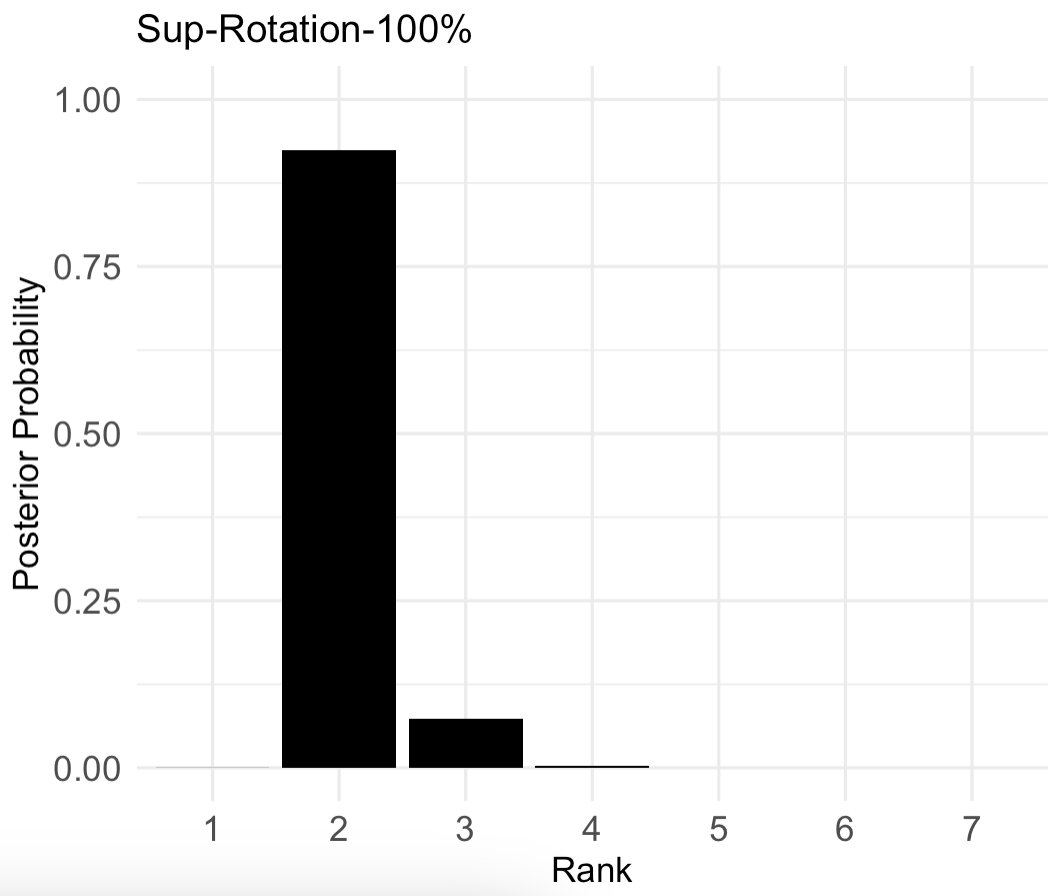}
\end{subfigure} & 
\begin{subfigure}[t]{0.3\textwidth}
    \includegraphics[width=\textwidth]{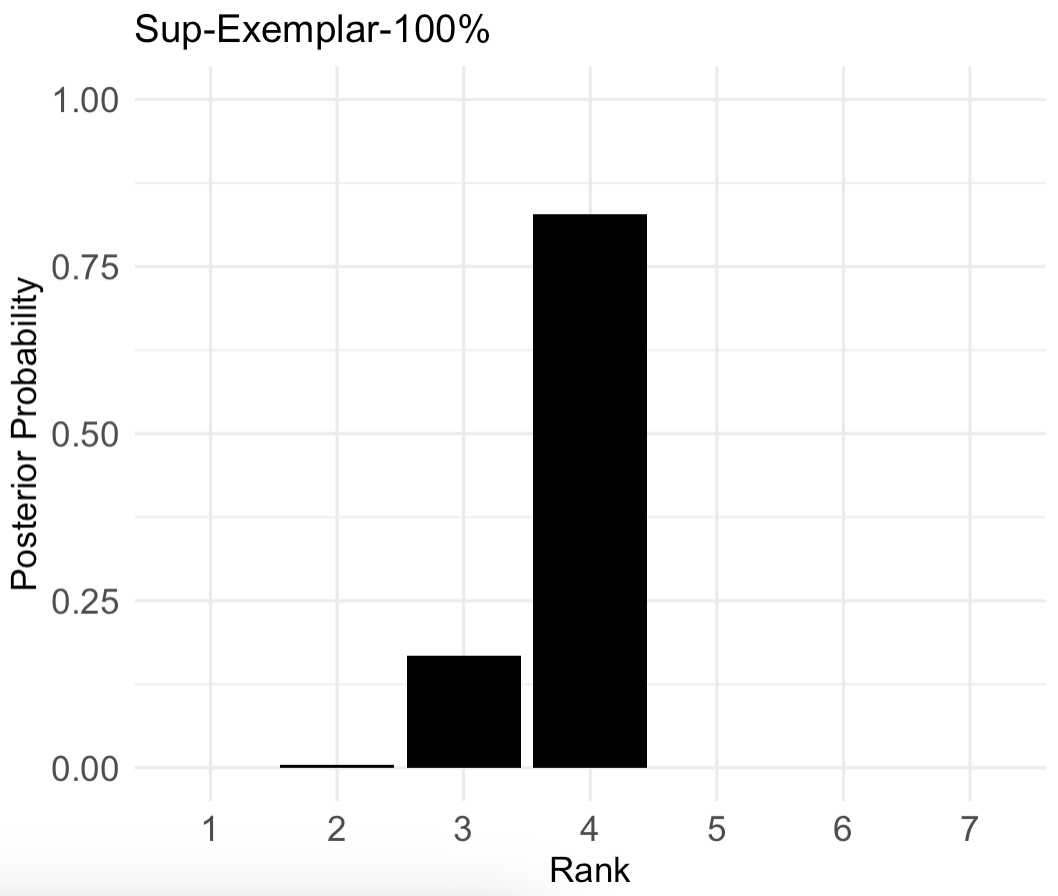}
\end{subfigure} &
\begin{subfigure}[t]{0.3\textwidth}
    \includegraphics[width=\textwidth]{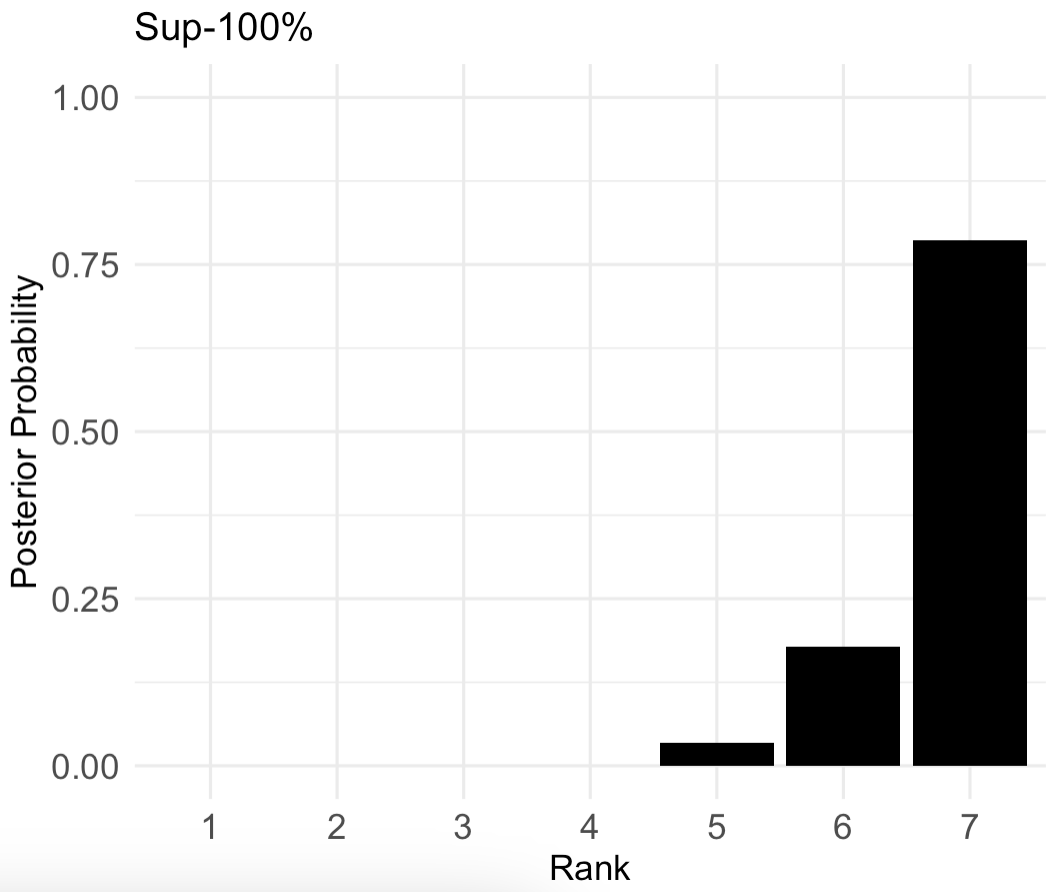}
\end{subfigure} \\ 
\begin{subfigure}[t]{0.3\textwidth}
    \includegraphics[width=\textwidth]{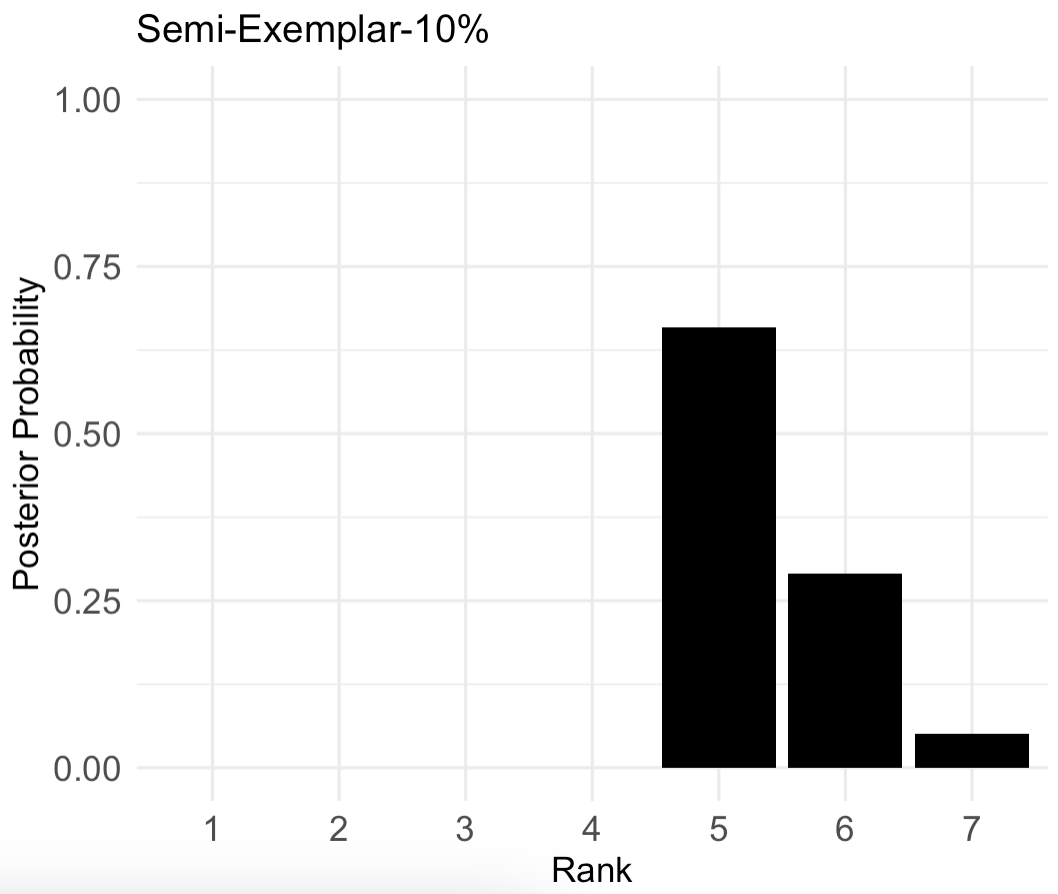}
\end{subfigure} &
\begin{subfigure}[t]{0.3\textwidth}
    \includegraphics[width=\textwidth]{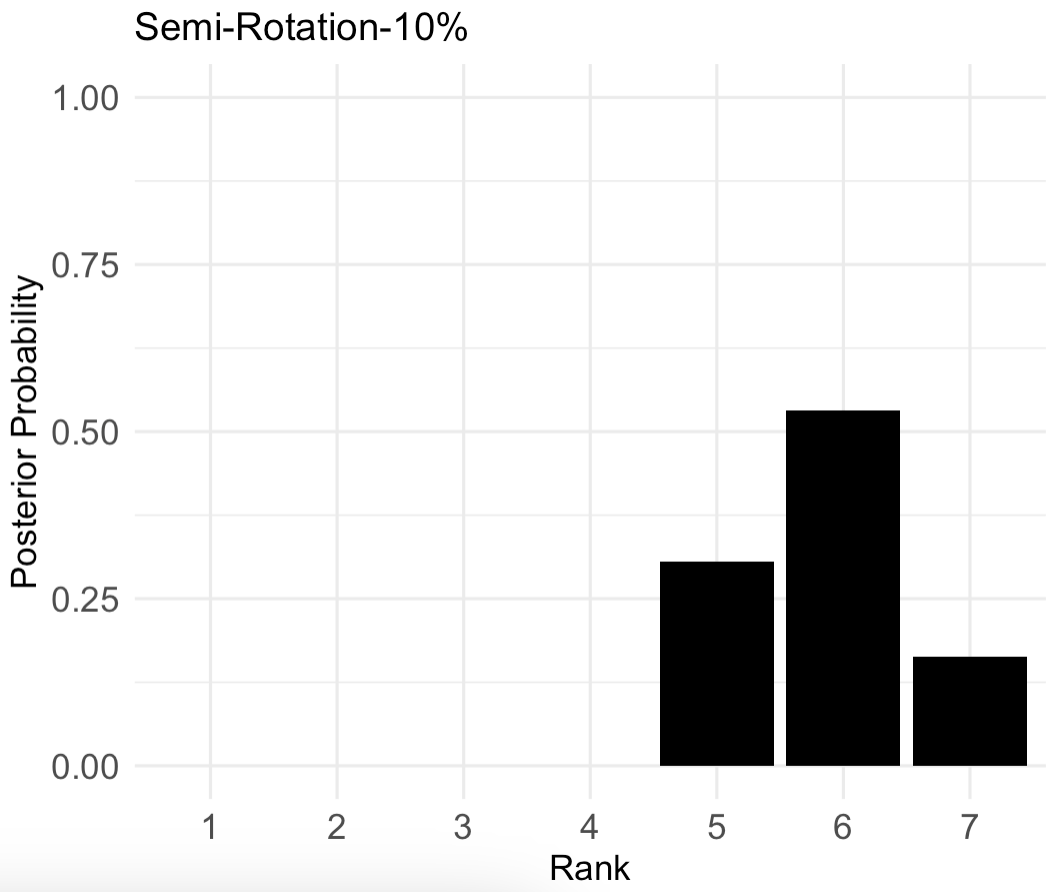}
\end{subfigure} & 
\begin{subfigure}[t]{0.3\textwidth}
    \includegraphics[width=\textwidth]{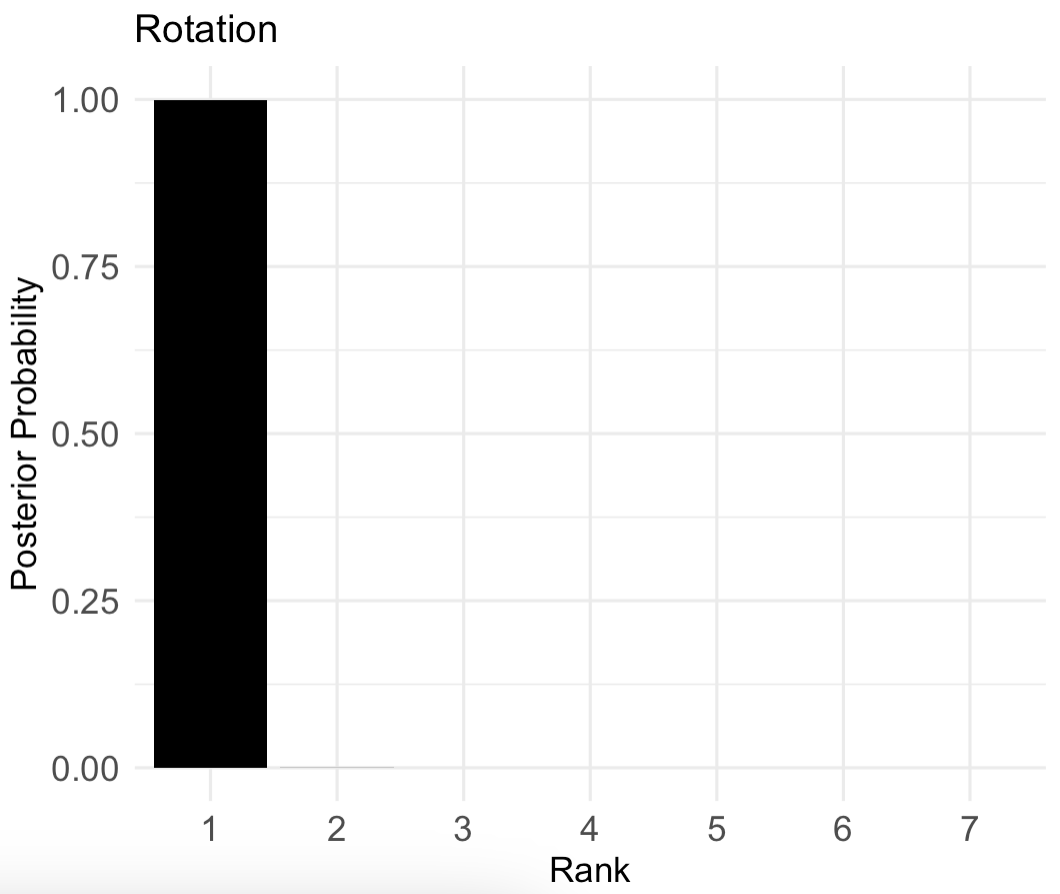}
\end{subfigure} \\
\end{tabular}
\caption{Posterior probabilities over model ranks for average accuracy weighted to favor structured image tasks using the Bayesian hierarchical model from \ref{sec:BHM} ($w_{Str} = 0.95$, $w_{Nat} = w_{Spe} = 0.025$).}
\label{fig:weighted-bayes-ranks}
\end{figure}

\begin{figure}
\setlength\tabcolsep{3pt}
\centering
\begin{tabular}{ccc}
\begin{subfigure}[t]{0.27\textwidth}
    \includegraphics[width=\textwidth]{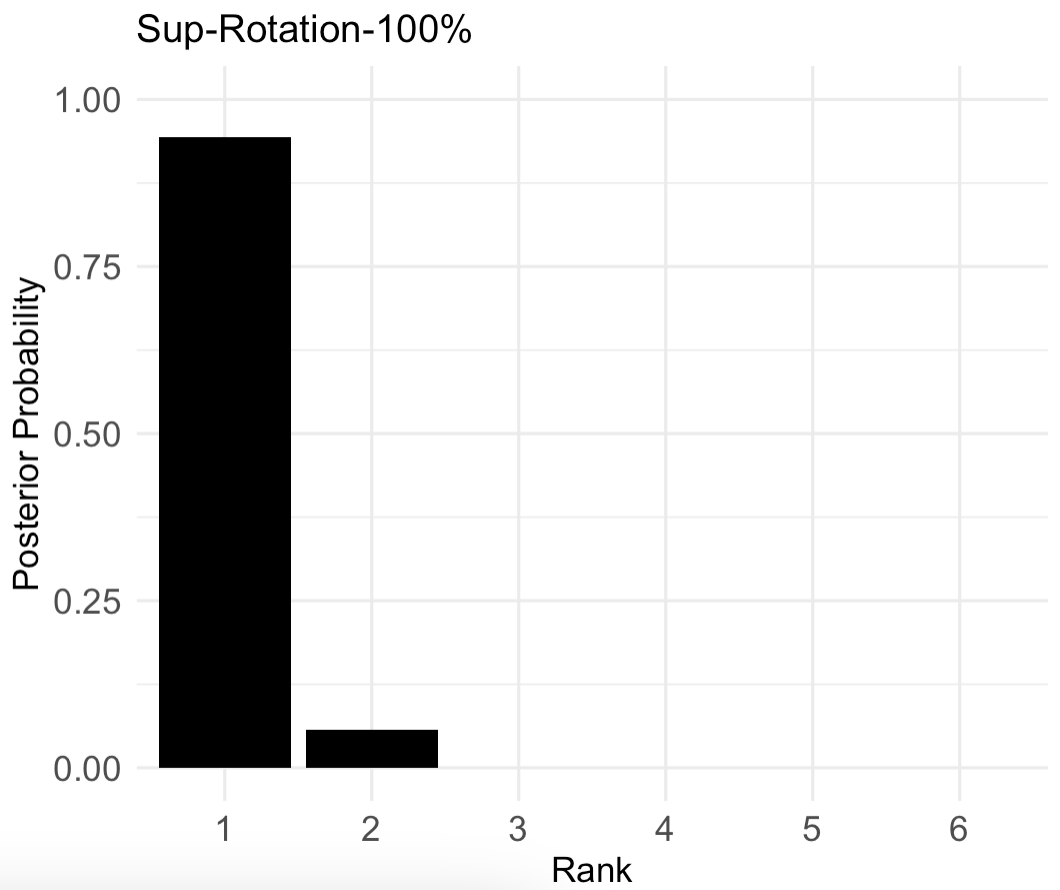}
\end{subfigure} & 
\begin{subfigure}[t]{0.27\textwidth}
    \includegraphics[width=\textwidth]{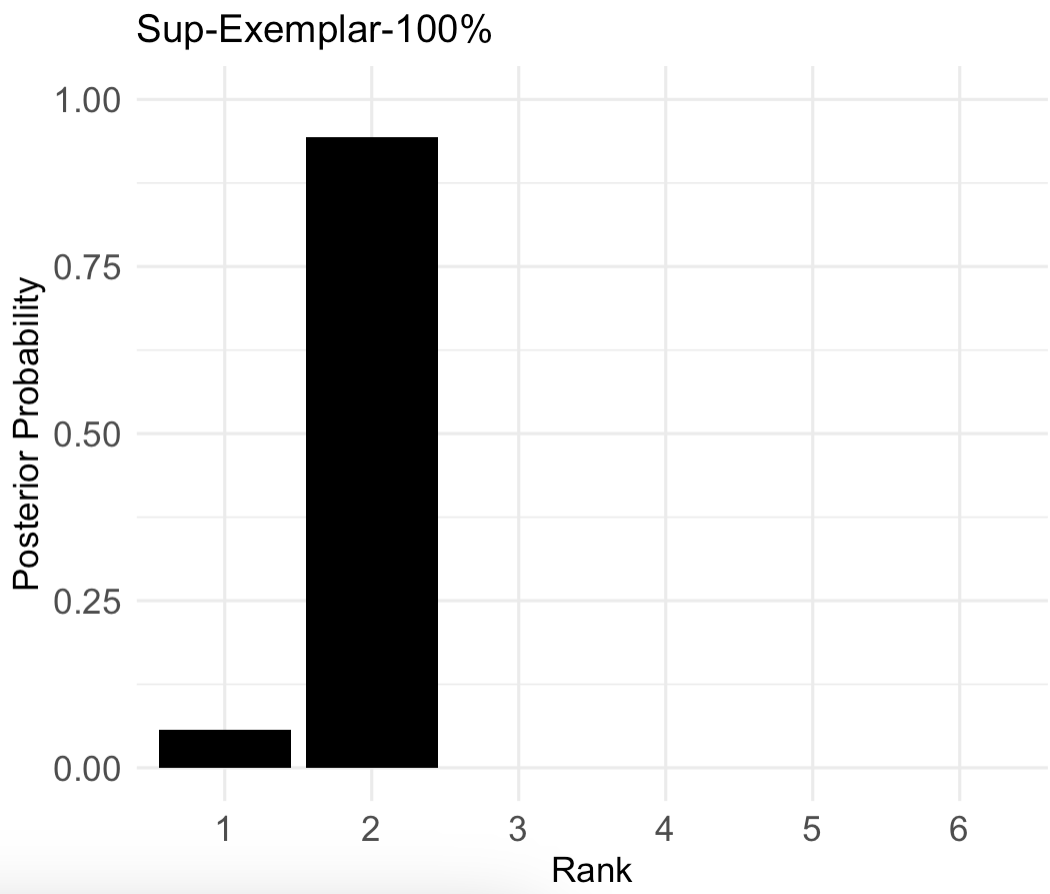}
\end{subfigure} &
\begin{subfigure}[t]{0.27\textwidth}
    \includegraphics[width=\textwidth]{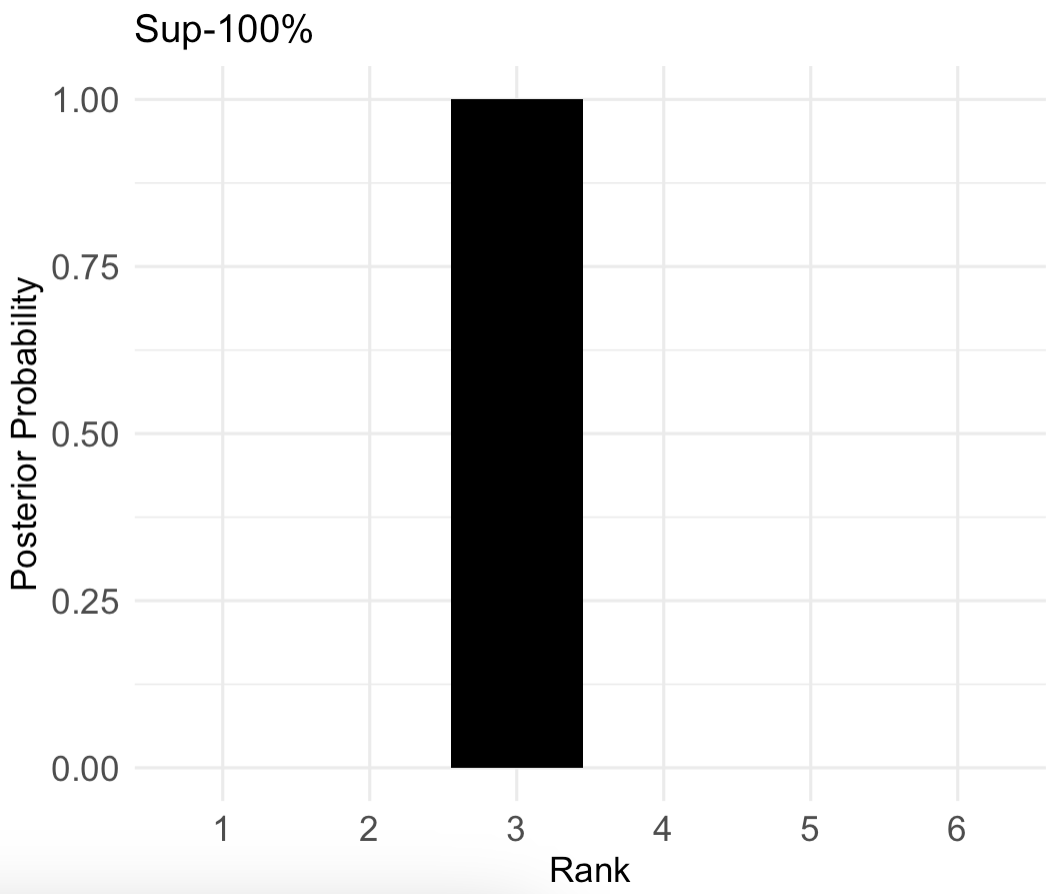}
\end{subfigure} \\ 
\begin{subfigure}[t]{0.27\textwidth}
    \includegraphics[width=\textwidth]{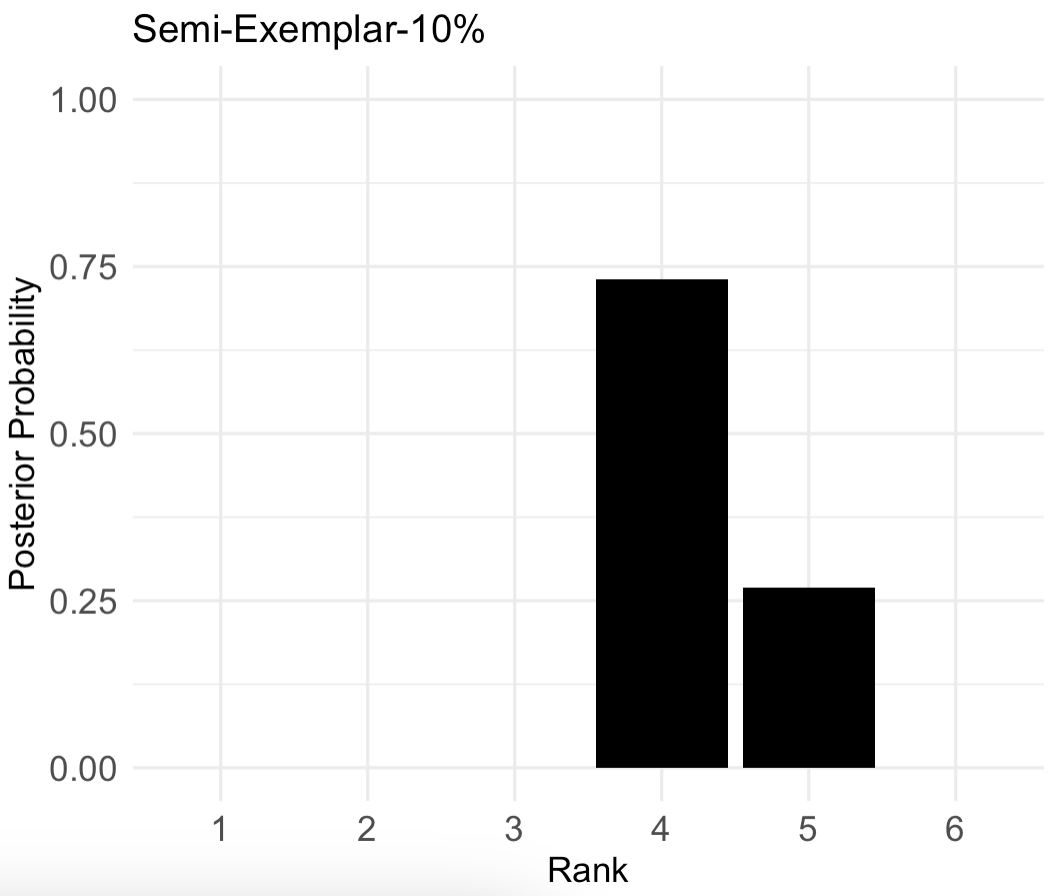}
\end{subfigure} &
\begin{subfigure}[t]{0.27\textwidth}
    \includegraphics[width=\textwidth]{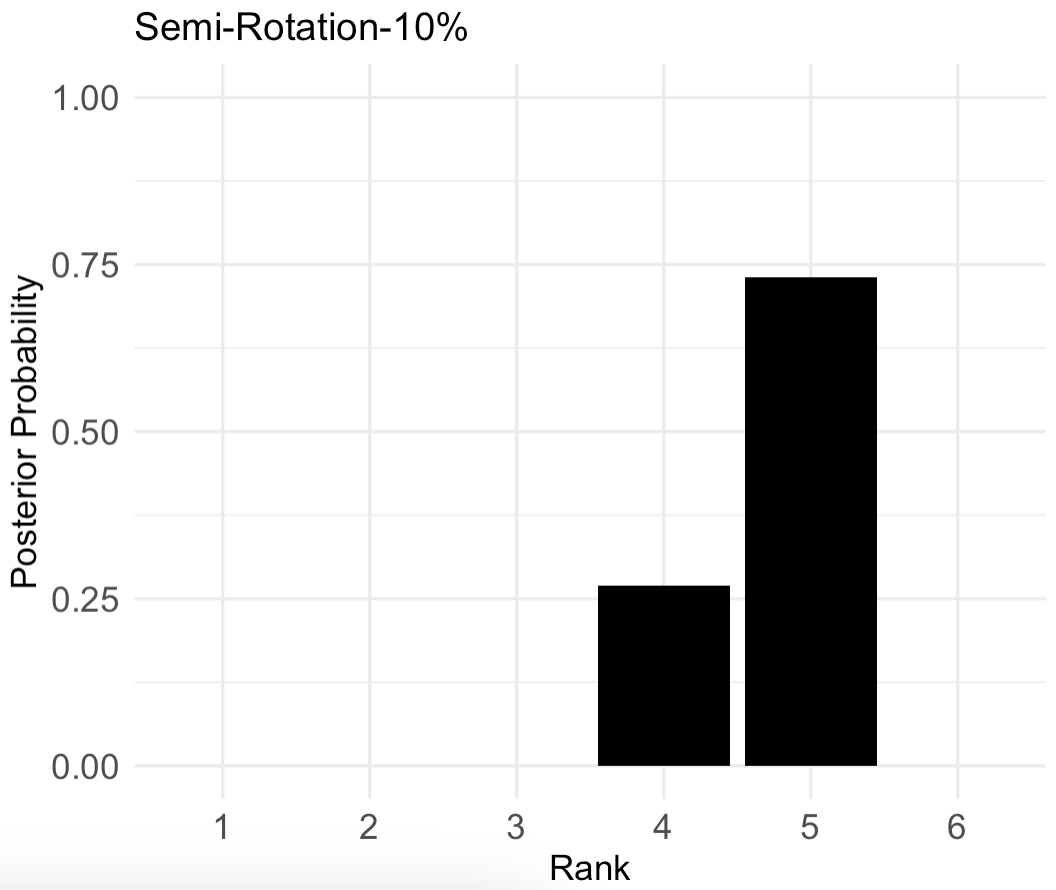}
\end{subfigure} & 
\begin{subfigure}[t]{0.27\textwidth}
    \includegraphics[width=\textwidth]{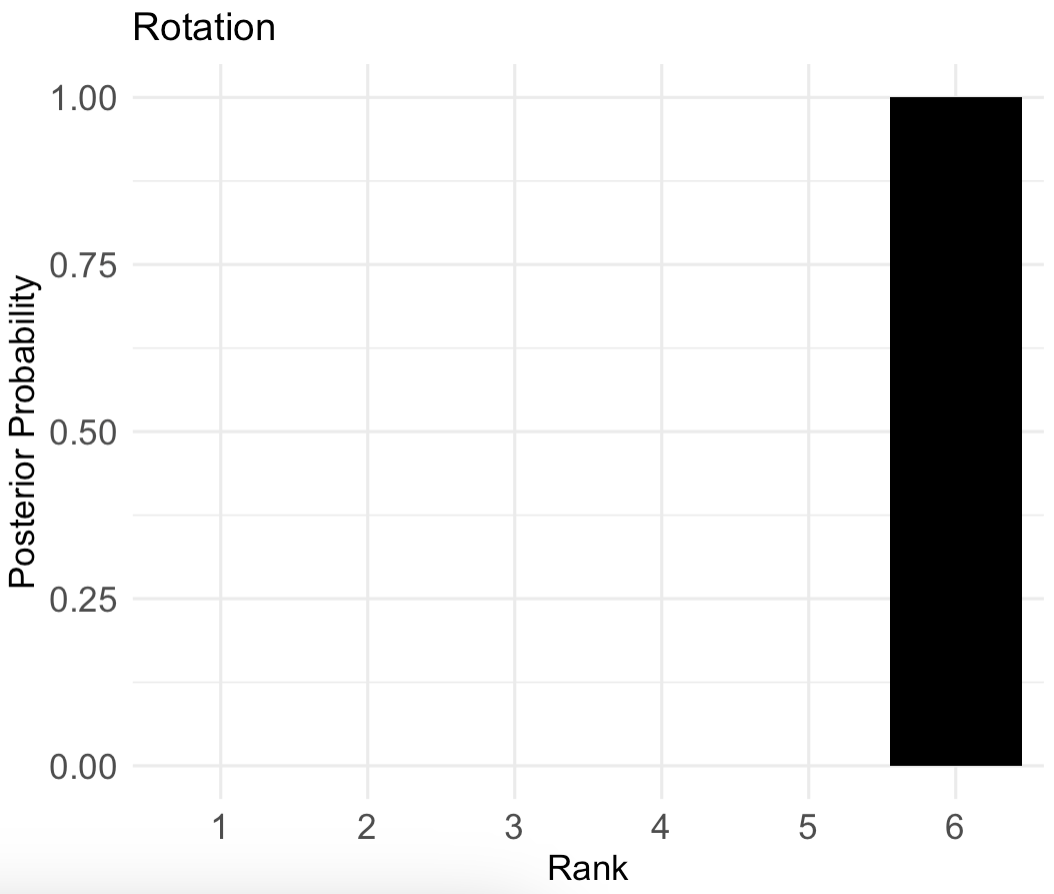}
\end{subfigure} \\
\end{tabular}
\caption{Posterior probabilities over model ranks for unweighted average accuracy using the Bayesian hierarchical model from \ref{sec:BHM}.}
\end{figure}
\clearpage

\end{document}